\documentclass[letterpaper]{article} 
\usepackage[preprint]{aaai23}  
\usepackage{times}  
\usepackage{helvet}  
\usepackage{courier}  
\usepackage[hyphens]{url}  
\usepackage{graphicx} 
\usepackage{natbib}  
\usepackage{caption} 
\usepackage{algorithm}
\usepackage[noend]{algpseudocode}

\usepackage{subfig}
\usepackage{array}

\usepackage{newfloat}
\usepackage{listings}
\DeclareCaptionStyle{ruled}{labelfont=normalfont,labelsep=colon,strut=off} 
\floatstyle{ruled}
\newfloat{listing}{tb}{lst}{}
\floatname{listing}{Listing}
\usepackage{transparent}

\usepackage{delimset}

\usepackage{amsthm}
\usepackage{amsmath}
\usepackage{amssymb}
\usepackage{amsfonts}
\usepackage{mathtools}
\usepackage{dsfont}

\usepackage{graphicx}

\usepackage{xcolor}
\definecolor{darkgreen}{rgb}{0,0.5,0}

\usepackage{cleveref}

\DeclareMathOperator*{\argmax}{arg\,max}

\DeclareMathOperator{\aset}{\mathcal{A}}

\DeclareMathOperator{\action}{a}

\DeclareMathOperator{\state}{s}

\DeclareMathOperator{\E}{\mathbb{E}} %

\newcommand{\CVaR}{\mathrm{CVaR}}
\newcommand{\VaR}{\mathrm{VaR}}

\DeclarePairedDelimiter\br{(}{)}
\DeclarePairedDelimiter\brs{[}{]}
\DeclarePairedDelimiter\brc{\{}{\}}

\newcommand{\indicator}[1]{\mathds{1}_{\brc{#1}}}

\newtheorem{theorem}{Theorem}
\newtheorem{definition}{Definition}

\def\S{{\mathcal S}}

\def\E{{\mathbb{E}}}

\def\int{{int}}
\def\ext{{ext}}

\newcommand{\cmnt}[1]{\ignorespaces}

\usepackage{multirow}

\title{Never Worse, Mostly Better:  \\ Stable Policy Improvement in Deep Reinforcement Learning}
\author{
    Pranav Khanna\textsuperscript{\rm3}\equalcontrib, Guy Tennenholtz\textsuperscript{\rm 2}\equalcontrib, Nadav Merlis\textsuperscript{\rm 2}\equalcontrib, Shie Mannor\textsuperscript{\rm 1,\rm2}\equalcontrib and Chen Tessler\textsuperscript{\rm 1}\equalcontrib
}
\affiliations{
    \textsuperscript{\rm 1} NVIDIA Research, \textsuperscript{\rm 2} Technion - Israel Institute of Technology, \textsuperscript{\rm3} Indian Institute of Technology, Indore
}

\usepackage{bibentry}

\begin{document}


\maketitle

\begin{abstract}
    In recent years, there has been significant progress in applying deep reinforcement learning (RL) for solving challenging problems across a wide variety of domains. Nevertheless, convergence of various methods has been shown to suffer from inconsistencies, due to algorithmic instability and variance, as well as stochasticity in the benchmark environments. Particularly, despite the fact that the agent's performance may be improving on average, it may abruptly deteriorate at late stages of training. In this work, we study methods for enhancing the agent's learning process, by providing conservative updates with respect to either the obtained history or a reference benchmark policy. Our method, termed EVEREST, obtains high confidence improvements via confidence bounds of a reference policy. Through extensive empirical analysis we demonstrate the benefit of our approach in terms of both performance and stabilization, with significant improvements in continuous control and Atari benchmarks.
\end{abstract}

\begin{figure*}[t!]
    \centering
    \includegraphics[width=\linewidth]{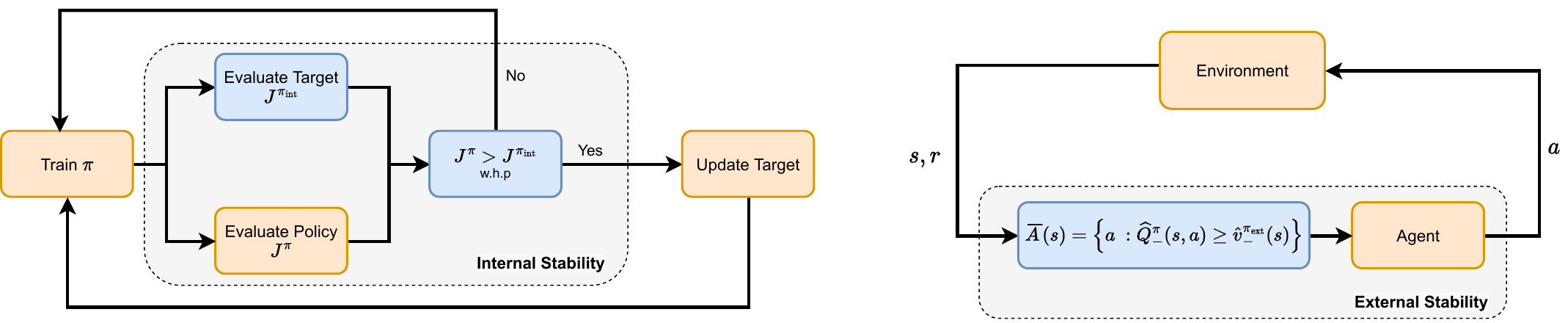}
    \caption{A diagram illustrating how EVEREST assures internal stability (left) and external stability (right). To attain internal stability, EVEREST requires the target policy to improve with high probability. Alternatively, external stability is achieved by limiting the agent to an admissible action set $\bar{\mathcal{A}}$ defined by actions that improve over a benchmark policy with high probability.}
    \label{fig: everest diagram}
\end{figure*}

\begin{figure*}[t!]
    \centering
    \setlength\extrarowheight{8pt}
    \begin{tabular}{cccc}
        \multicolumn{2}{c|}{\textbf{Hopper (Internal Stability)}} & \multicolumn{2}{c}{\textbf{PacMan (External Stability)}} \\
        \multicolumn{2}{c|}{\hspace{0.8cm}\textbf{Oblivious} \hspace{2.7cm} \textbf{EVEREST}} & \multicolumn{2}{c}{\hspace{0.6cm}\textbf{Return} \hspace{2.9cm} \textbf{Regret}}\\
        \multicolumn{2}{c|}{
            \multirow{-3.9}{*}{
                \subfloat[\label{fig:td3 issues}]{
                    \includegraphics[width=0.23\linewidth]{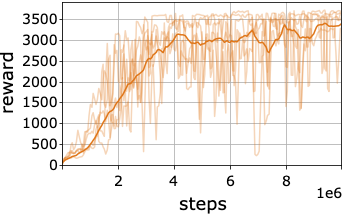}
                    \includegraphics[width=0.23\linewidth]{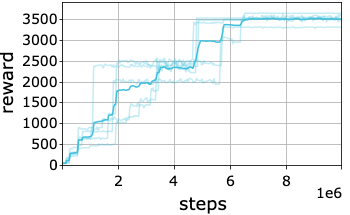}
                }
            }
            
        } & \multicolumn{2}{c}{
            \subfloat[\label{fig: freeway safe learning}]{
                \includegraphics[width=0.23\linewidth]{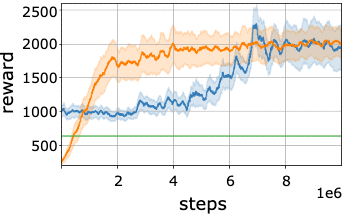}
                \includegraphics[width=0.23\linewidth]{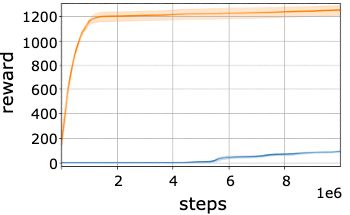}
            }
        } \\
        \multicolumn{4}{c}{
            \includegraphics[width=0.5\linewidth,trim={0.1cm 0 0 0},clip]{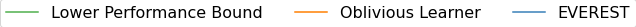}
        }
    \end{tabular}
    \caption{\textbf{(a) Internal instability:} Performance plots of TD3 \citep{fujimoto2018addressing}. Transparent lines (not smoothed across time) represent individual seeds, whereas opaque lines represent the average smoothed mean over seeds. The left plot presents the standard training process (oblivious learner), whereas the right plot represents a conservative update process that ensures improvement with high probability. While the overall trend is improving, at various points throughout training, the oblivious learner exhibits bad performance. \textbf{(b) External instability:} Performance and regret comparison between EVEREST and an oblivious learner. The goal of EVEREST, a safe and stable learner, throughout the entire learning process, is to perform at least as good as the lower confidence bound estimate for the benchmark. The regret measures the accumulated number of sub-optimal episodes, measured with respect to the performance of the benchmark policy.}\label{fig: intro}
\end{figure*}

\section{Introduction}
Instability of reinforcement learning (RL) methods has been a longstanding problem within the research community \cite{henderson2018deep}. While most research in deep RL focuses on metrics such as `best performance obtained during training' or `final converged performance', the \textit{behavior} of the agent \textit{during} training is often ignored. Such agents may be highly unstable and present significant safety concerns in real-world deployments.

This work focuses on improving the behavior exhibited during training -- \textbf{seeking stable and always improving agents}. We aim to measure the reliability of each individual training run, and suggest both \textit{internal} and \textit{external} stabilization methods to resolve these issues (see \Cref{fig: everest diagram}).


Internal stability is concerned with the agent itself and its own historical behavior. Ideally, the agent should be monotonously improving \citep{viering2019open,bousquet2022monotone}. Nevertheless, the learning process of deep RL agents is characterized by frequent instabilities in performance (\Cref{fig:td3 issues}). This is often unnoticed, as smoothed learning curves give an illusion of stability. Although the general trend is often improving on average, halting the agent at a random point may result in arbitrarily poor performance. Internal stability aims to provide learners with high probability improvement guarantees.

Alternatively, to complement internal stability, a learner is often able to access an external benchmark policy. Instead of attempting to imitate this policy, one may utilize it as a stabilizing benchmark, requiring the agent to always perform better. Achieving external stability is related to safety requirements, as agents learning in the real world should never underperform particular benchmark policies (e.g., a human driver in a vehicle, or a classic controller for a quadcopter \citep{argentim2013pid}). A comparison of the expected behaviors of a regular agent and an externally stabilizing agent is presented in \Cref{fig: freeway safe learning}. Indeed, throughout the training process, the learner performs at least as well as the benchmark.

For both internal and external stability, we use statistical tests to determine deterioration. To ensure internal stability, we update a proposed target network only when it has improved with high probability. To ensure external stability, we limit the learner to actions that improve upon a baseline with high probability, or otherwise switch to the benchmark policy.

Our contributions are as follows: (1) We present novel internal and external stabilization methods for deep RL algorithms. (2) We motivate our approach theoretically and conduct empirical analysis of instabilities in concurrent deep RL methods. (3) We show through extensive experiments that our learning schemes result in increased stability. Empirical evidence shows that the agent is less likely to behave sub-optimally. Our results suggest that our proposed methods result in improved behavior both in terms of stability as well as overall performance.

\section{Preliminaries}\label{sec: background}

We consider an infinite-horizon Markov Decision Process \citep[MDP]{puterman1994markov} defined by the tuple $(\S, \mathcal{A}, \mathcal{R}, \mathcal{P}, \gamma)$, where $\S$ are the states, $\mathcal{A}$ the actions, $\mathcal{R} : \S \to \mathbb{R}$ the reward function, $\mathcal{P} : \S \times \mathcal{A} \to \S$ is the transition kernel and $\gamma \in [0, 1)$ is the discount factor that governs whether the agent is myopic or forward looking.

A policy $\pi : \S \to \Delta_\mathcal{A}$ defines, for each state, a probability distribution over actions. In addition to the policy, the value function $v^\pi (s) : \S \to \mathbb{R} = \E^\pi [\sum_t \gamma^t r_t | s_0 = s]$ is the expected reward-to-go of the policy $\pi$ starting from state $\state$, and the quality function $Q^\pi (s, a) : \S \times \mathcal{A} \to \mathbb{R} = \E [ \sum_t \gamma_t r_t | s_0 = s, a_0 = a]$ is the utility of initially playing action $a$ and then acting based on policy $\pi$ afterwards. In addition, we define the expected performance of a policy $\pi$ by $J^\pi = \E_{\state \sim \rho} v^\pi (\state)$, where $\rho$ is the initial state distribution. Finally, we denote the policy at episode $k$ by $\pi_k$ and its parameterized representation as $\pi_{\theta_k}$, where $\theta_k$ are the parameters at episode $k$.

\paragraph{Statistical Testing.}

In this work, we will utilize statistical tests in order to control various aspects of the learning process. Given a random variable (RV) $X$, we denote the sample mean by $\hat X$, the sample variance by $\hat \sigma_X$ and the number of samples by $n_X$. The lower confidence bound is then defined as $\hat X_{-} = \hat X - \hat \sigma_{X} \cdot C$, where $C$ is a constant controlling the tightness of the confidence bound. A statistical test is used to perform quantitative decisions about a process. 

A statistical test involves the null hypothesis, e.g., ${X \le Y}$. Assuming a Gaussian distribution of the estimates, this test can be performed using Welch's one-tailed t-test \citep{welch1947generalization}. The $t$-statistic is then defined as
\begin{equation*}
    t = \left(\hat{X} - \hat{Y}\right) / \left(\sqrt{{\hat{\sigma}^2_{X}}/{n_{X}} + {\hat{\sigma}^2_{Y}}/{n_{Y}}}\right) \,.
\end{equation*}
This statistic enables us to find the $p$-value, which represents the confidence in the null hypothesis. Under the Gaussian assumption, the $p$-value can be directly calculated by $p=1-\Phi(t)$, where $\Phi(x)$ is the cumulative distribution function of a standard normal random variable. Hence, when $p > 1 - \delta$ we have a high enough confidence that the null hypothesis is incorrect, meaning, in the example above, that $Y$ is actually greater than $X$.

\begin{figure*}[t!]
    \centering
    \begin{tabular}{cc}
    \textbf{Median Performance across Runs} & \textbf{Lower CVaR on Differences} \\ \vspace{0.2cm}
    \subfloat[\label{fig: median}]{\includegraphics[width=0.4\linewidth]{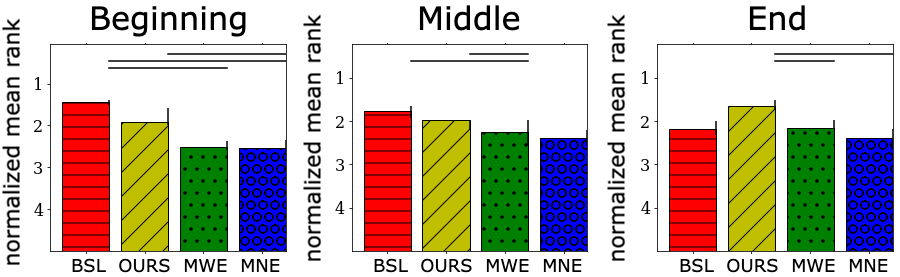}} & 
\multirow{-7.2}{*}{\subfloat[\label{fig: comparisons}]{\includegraphics[width=0.55\linewidth,,height=0.325\linewidth]{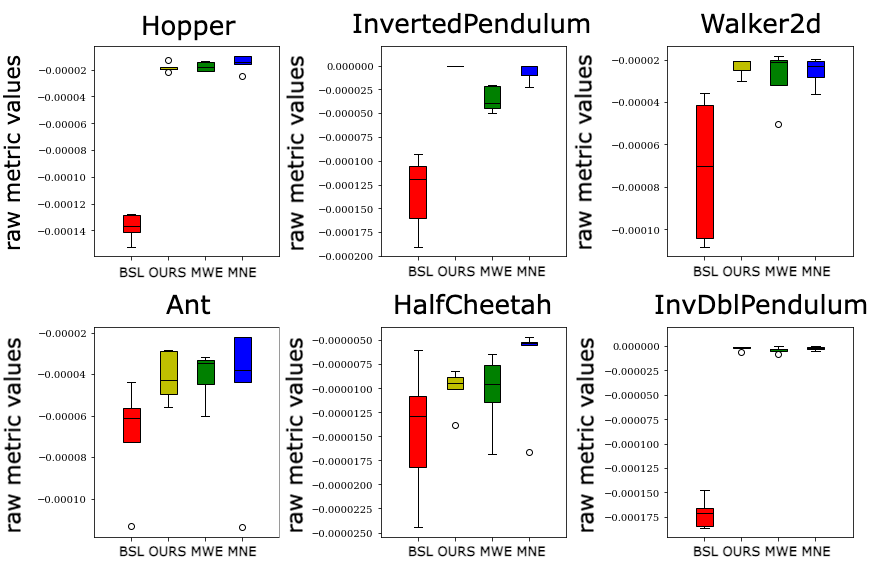}}} \\
    \textbf{IQR across Time} & \vspace{0.2cm}\\
    \subfloat[\label{fig: iqr}]{\includegraphics[width=0.4\linewidth]{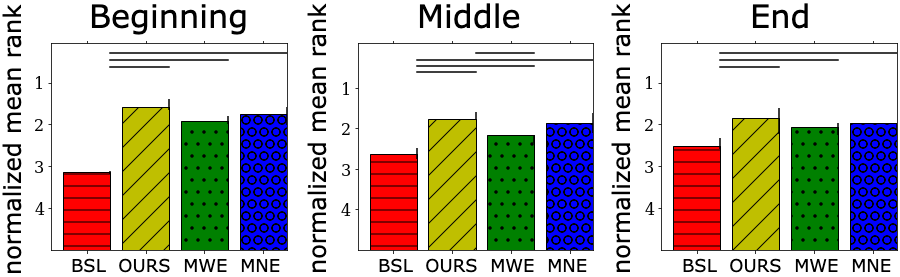}} & 
    \end{tabular}
    
    \caption{\textbf{Internal stability:} Reliability metrics and median performance for BSL (Oblivious TD3 Baseline), Ours (EVEREST), MWE (Max w/ reevaluation), and MNE (Max w/o reevaluation). Error bars are 95\% bootstrap confidence intervals (\# bootstraps = 1,000). Significant pairwise differences in ranking between pairs of algorithms are indicated by black horizontal lines above the colored bars and a higher rank is better. In (a) we present the aggregate median performance across environments and seeds, measured over 3 phases of training. In (b) we present the maximal average draw-down of the performance during training, measured as the CVaR over the 25th percentile. Better reliability is indicated by more positive values. Finally, (c) measures the intraquartile range, the dispersion over 25th to 75th percentiles. This metric is also reported as an aggregate across experiments.}
    \label{fig: comparisons2}
\end{figure*}

\section{EVEREST}

In this section we define internal and external stability of RL agents, and propose methods to achieve them. We present our approach, "nEVER woRsE moStly beTter" (EVEREST), an off-policy method which alternates between learning and evaluation phases. During training, the agent collects data using an online policy, and is periodically paused for evaluation. Internal stability is ensured through high probability updates of a target network, while external stability is achieved through high probability constraints of an admissible action set. Pseudocode for EVEREST is shown in \Cref{alg:cap} along with a block diagram in \Cref{fig: everest diagram}. In what follows, we define and analyze internal and external stability separately.



\paragraph{Internal Stability.} We begin by defining internal stability of agents, for which every training episode improves upon the previous with high probability.

\begin{definition}[Internal Stability]
    We say that an RL algorithm is $\delta$-internally-stable if, with probability at least $1-\delta$, for every episode $k \in \brk[c]*{1, 2, \hdots}$, $J^{\pi_k} \geq J^{\pi_{k-1}}$. 
\end{definition}

A common approach in off-policy learning is to use target networks to stabilize learning \citep{mnih2015human,haarnoja2018soft,fujimoto2018addressing}, by slowly following the policy performance, often using Polyak-Rupert averaging \citep{polyak1990new}. To achieve internal stability, we propose to condition this update according to the likelihood of improvement. Particularly, at every fixed interval, the learner proposes a new target network, which is updated if it improves upon the current target network with high probability.

Formally, given performance estimates $\hat J$ with standard deviation $\hat \sigma$ that were obtained using $n$ samples from $\pi_\theta$ and the target $\pi_{\theta_{\text{int}}}$, we formulate the null hypothesis as ${J^{\pi_\theta} \leq J^{\pi_{\theta_{\text{int}}}}}$. Following \Cref{sec: background}, the target network is updated when the null hypothesis is rejected. That is,
\begin{equation}\label{eqn: t test internal stability}
    \frac{\hat{J}^{\pi_{\theta}} - \hat{J}^{\pi_{\theta_{\text{int}}}}}  {  \sqrt{{\hat{\sigma}^2_{J^{\pi_\theta}}}/{n_{\pi_{\theta}}} + {\hat{\sigma}^2_{J^{\pi_{\theta_{\text{int}}}}}}/{n_{\pi_{\theta_{\text{int}}}}}}} > \Phi^{-1}(1-\delta) \,.
\end{equation}

\paragraph{External Stability.} Next, we define external stability with respect to a benchmark policy, which ensures that during training, the current learned policy is at least as good as a benchmark, with high probability.

\begin{definition}[External Stability]
    We say that an RL algorithm is $\delta$-externally-stable w.r.t. a benchmark $\pi_{\text{\ext}}$ if, with probability at least $1-\delta$, for every episode $k \in \brk[c]*{1, 2, \hdots}$, $J^{\pi_k} \geq J^{\pi_{\text{ext}}}$.
\end{definition}

To achieve external stability we first show how benchmark policies can be used to improve RL agents. The following theorem states that given a set of policies $\brk[c]*{\pi^{(i)}}_{i=1}^M$, a per-state value maximizing policy, denoted by $\bar{\pi}$, achieves higher performance than any individual policy in the set. The proof is provided in the supplementary material for both the discounted and finite horizon settings.

\begin{table*}[t!]
    \small
    \setlength\extrarowheight{4pt}
    \centering
    \begin{tabular}{c|c|c|c|c|c|c}
         & \textbf{Hopper} & \textbf{InvertedPendulum} & \textbf{InvertedDoublePendulum} & \textbf{Walker2d} & \textbf{HalfCheetah} & \textbf{Ant} \\ \hline\hline
        \textbf{EVEREST (ours)} & $\mathbf{3509 \pm 12}$ & $\mathbf{1000 \pm 0}$ & $\mathbf{9359 \pm 0}$ & $\mathbf{4418 \pm 35}$ & $\mathbf{10397 \pm 66}$ & $4512 \pm 63$ \\ \hline
        \textbf{Max w/ reevaluation} & $\mathbf{3533 \pm 17}$ & $\mathbf{1000 \pm 0}$ & $\mathbf{9355 \pm 0}$ & $4027 \pm 50$ & $9242 \pm 29$ & $4598 \pm 64$ \\ \hline
        \textbf{Max w/o reevaluation} & $\mathbf{3538 \pm 23}$ & $990 \pm 3$ & $\mathbf{9358 \pm 0}$ & $4106 \pm 31$ & $9087 \pm 110$ & $4125 \pm 46$ \\ \hline
        \textbf{Oblivious} & $3424 \pm 249$ & $971 \pm 54$ & $8961 \pm 584$ & $4037 \pm 278$ & $7968 \pm 80$ & $\mathbf{4892 \pm 203}$
    \end{tabular}
    \caption{\textbf{Internal stability:} Reliability of the final converged performance. We measure the mean and standard deviation of the final phase of the learning process. To do so, for each seed, we record the final $10$ epochs. Each seed is evaluated for $100$ episodes. The mean performance for each seed is the average over all evaluations (mean of means) and the standard deviation is the standard deviation across means. To aggregate across seeds, we report the average values taken over seeds. While the goal is higher final performance, a lower standard deviation means lower variation in the final phase of the learning process and thus higher reliability.}
    \label{tab: process reliability}
\end{table*}

\begin{theorem}\label{thm: improvement theorem}
    Let $\brk[c]*{\pi^{(i)}}_{i=1}^M$ and define $\bar{\pi}$ such that for all ${s \in \S}$, $\bar{\pi}(s) \in \arg\max_{i \in [M]} v^{\pi^{(i)}}(s)$. Then,
    \begin{align*}
    v^{\bar{\pi}}(s) \geq \max_{i \in [M]} v^{\pi^{(i)}}(s), \forall s \in \S.
    \end{align*}
\end{theorem}

Next, we leverage the result of \Cref{thm: improvement theorem} for external stability. Particularly, we assume the agent has access to a benchmark policy $\pi_{\text{ext}}$, and a lower performance bound of $J^{\pi_{\text{ext}}}$, which we denote by $J^{\pi_{\text{ext}}}_{-}$. To achieve external stability, we utilize the result of \Cref{thm: improvement theorem} for the case of two policies -- the learner $\pi$ and the benchmark policy $\pi_\text{ext}$. To construct the mixture policy $\bar{\pi}$, we utilize an action elimination scheme; namely, at each state, the learner determines and constrains itself to the set of actions that improve its performance with high probability. The admissible action set is defined by
\begin{equation}\label{eqn: admissible action set}
    \bar{\mathcal{A}} (s) = \{ \action \in \mathcal{A} : Q^\pi_{-} (\state, \action) \geq v^{\pi_{\text{ext}}}_{-} (\state) \} \,,
\end{equation}
where $Q^\pi_{-}$ is a lower performance bound of $Q^\pi$. To achieve lower performance bounds we use an ensemble of networks and a bootstrap scheme \citep{osband2016deep}. Following \Cref{thm: improvement theorem}, any policy that is defined over $\Delta_{\bar{\mathcal{A}}}$ will achieve external stability. In this work, we consider the standard $\epsilon-$greedy policy.

In the next section, we conduct experiments for both internal and external stability. We show that our methods drastically improve the stability and performance of the agent compared to non-stabilizing baselines, and measure the benefit of our methods using recently proposed dispersion and risk metrics \citep{rl_reliability_metrics}.






\algrenewcommand\algorithmicensure{\textbf{Initialize}}
\begin{algorithm}[t!]
    \caption{nEVEr woRsE moStly beTter (EVEREST)}\label{alg:cap}
    \begin{algorithmic}
        \Require Ensemble size $N$, $\hat v^{\pi_{\text{ext}}}_{-}$, target update interval $T$, confidence $\delta$ and environment $env$
        \Ensure Ensemble $\{Q_{_i}\}_{i=1}^N$ and agent $\pi_\theta$, replay buffer $\mathcal{D}$, target network policy $\theta_{\text{int}}$.
        \State $\state_1 \gets env.reset()$
        \For{$t = 1, 2, \ldots$}
            \State $\bar{\mathcal{A}} = \{ \action \in \mathcal{A} : \hat Q^\pi_{-} (\state, \action) \geq \hat v^{\pi_{\text{ext}}}_{-} (\state) \}$ \Comment{\cref{eqn: admissible action set}}
            \If{$\bar{\mathcal{A}}$ is not empty}
                \State $\action_t \gets \text{explore \& exploit} (\bar{\mathcal{A}}, \{Q_i\}_{i=1}^N, \pi)$
            \Else
                \State $\action_t \sim \pi_{\text{ext}} (\state)$
            \EndIf
            \State $\state_{t+1}, r_t \gets env.step(\action_t)$
            \State $\mathcal{D}.insert(\state_t, \action_t, r_t, \state_{t+1})$
            \State Update $\pi_\theta, \{Q_{_i} \}_{i=1}^N$ using replay buffer samples
            \If{$t \% T == 0$ and $P(J^{\pi_\theta} \geq J^{\pi_{\theta_{\text{int}}}}) \geq 1 - \delta$}
                \State $\theta_{\text{int}} \gets \theta$ \Comment{\cref{eqn: t test internal stability}}
            \EndIf
        \EndFor
    \end{algorithmic}
\end{algorithm}

\section{Experiments}

In this section we analyze our approach by focusing on three questions: (1) Do contemporary methods suffer from instability? (2) Does EVEREST empirically improve stability of these methods? (3) Does EVEREST improve the performance of these methods? In what follows we answer all three of these questions affirmitively, through extensive analysis in continuous control and Atari benchmarks.

\subsection{Implementation Details}

We begin with a brief overview of the learning process, followed by a thorough empirical analysis of EVEREST. As internal and external stability are orthogonal, we analyze each independently. For internal stability, we analyze the Twin-Delayed DDPG \citep[TD3]{fujimoto2018addressing} algorithm, for which we replace the target policy $\pi_{\theta_{\text{int}}}$ update scheme, as shown in \Cref{alg:cap}. For external stability we consider both Soft Actor Critic \citep[SAC]{haarnoja2018soft} and Double DQN \citep[DDQN]{van2016deep}. To estimate the admissible action set (\Cref{eqn: admissible action set}), we use an ensemble of $Q-$critics.

For all the experiments, we run 5 different random seeds. For internal stability, EVEREST performs policy evaluation as an intrinsic property of the algorithm. Hence, for fair comparison, these policy evaluation steps are considered part of the training budget.

As we analyze the behavior of the training process, we need to ensure that any observed performance degradation is due to the policy optimization and not due to the sampling techniques. Hence, throughout training, each evaluation is performed over 100 episodes. In contrast to the evaluations EVEREST performs, these evaluations are for reporting only and are thus not considered part of the training budget.

\begin{figure}[t]
    \centering
    \hspace{1cm}\textbf{Oblivious Learner} \hfill \textbf{EVEREST}\hspace{1.15cm}\vspace{0.1cm} \\
    \hspace{0.4cm} InvertedDoublePendulum \hspace{0.45cm} InvertedDoublePendulum \\
    \includegraphics[width=0.49\linewidth]{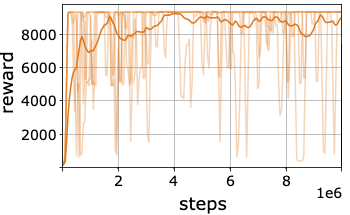} \includegraphics[width=0.49\linewidth]{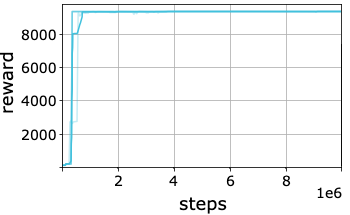} \\ \vspace{0.1cm}
    \hspace{0.5cm} Ant \hspace{3.5cm} Ant \\
    \includegraphics[width=0.49\linewidth]{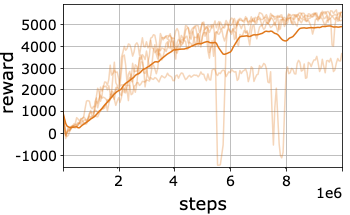} \includegraphics[width=0.49\linewidth]{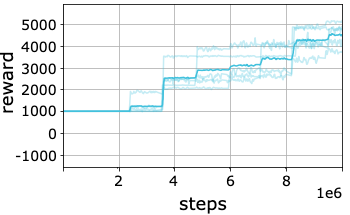}
    \caption{\textbf{Internal Stability:} Training plots comparing an oblivious learner (the baseline TD3 algorithm) with EVEREST. As is seen across all tested domains, EVEREST exhibits increased \textit{per seed} stability, ensuring a monotonously increasing performance (w.h.p.).}
    \label{fig: partial internal stability}
\end{figure}

\subsection{Internal Stability}\label{sec: exp internal}

We begin by analyzing internal stability. For this, we consider three parameters: (1) how often we evaluate the online policy, (2) how many episodes are used for evaluation, and (3) the confidence level $\delta$ required for policy improvement.

The confidence level and the number of evaluations are strongly connected, as additional evaluations will increase the confidence. On the other hand, the evaluation frequency controls how fast the process converges. For our experiments, we performed an initial hyperparameter tuning phase on the Hopper domain. The same hyperparameters were then used for all other tasks.


We plot both the individual seeds (transparent curves) and the average results across seeds (opaque curves). As opposed to previous works, \textbf{we do not smooth the per-seed graphs} using a moving average procedure, thus providing better insight to the experiment variance and stability.

\begin{table*}
    \centering
    \resizebox{0.85\linewidth}{!}{
    \small
    \setlength\extrarowheight{4pt}
    \centering
    \begin{tabular}{c|c|c|c|c|c|c|c}
        \multicolumn{2}{c|}{} & \multicolumn{2}{c|}{\textbf{Ant}} & \multicolumn{2}{c|}{\textbf{HalfCheetah}} & \multicolumn{2}{c}{\textbf{Humanoid}} \\ \hline\hline
        \multirow{2}{*}{\textbf{Benchmark} ($\pi_{\text{ext}}$)} &  & Med & Good & Med & Good & Med & Good \\
        \cline{2-8}
         & $\hat J^{\pi_{\text{ext}}}_{-}$ & $1721$ & $3328$ & $5406$ & $8004$ & $714$ & $915$ \\ \hline\hline
        \multirow{2}{*}{\textbf{EVEREST (ours)}} & Return $\uparrow$ & $\mathbf{7110 \pm 118}$ & $\mathbf{7006 \pm 56}$ & $\mathbf{16242 \pm 215}$ & $15316 \pm 210$ & $6087 \pm 44$ & $6137 \pm 44$ \\
        \cline{2-8}
         & Regret $\downarrow$ & $\mathbf{163 \pm 3}$ & $\mathbf{82 \pm 3}$ & $\mathbf{96 \pm 3}$ & $\mathbf{68 \pm 3}$ & $\mathbf{758 \pm 6}$ & $\mathbf{450 \pm 6}$ \\
        \hline
        \multirow{2}{*}{\textbf{Oblivious}} & Return $\uparrow$ & \multicolumn{2}{c|}{$5007 \pm 182$} & \multicolumn{2}{c|}{$15148 \pm 332$} & \multicolumn{2}{c}{$\mathbf{6454 \pm 446}$} \\
        \cline{2-8}
         & Regret $\downarrow$ & $1236 \pm 34$ & $1574 \pm 33$ & $182 \pm 10$ & $334 \pm 7$ & $2524 \pm 116$ & $3021 \pm 159$ \\
    \end{tabular}
    }
    \caption{\textbf{External stability in MuJoCo:} We compare EVEREST with an oblivious learner. We compare against medium and good benchmark agents. $\hat J^{\pi_{\text{ext}}}_{-}$ is the measured lower-confidence bound of the benchmark. We measure the return (higher is better) and regret (lower is better) of the agents. Line plots are presented in the supplementary material. In addition to improving stability, EVEREST is capable of outperforming the standard learning procedure by $40\%$ on Ant.}
    \label{tab: process safety mujoco}
\end{table*}

\begin{table*}
    \resizebox{\linewidth}{!}{
    \small
    \setlength\extrarowheight{4pt}
    \centering
    \begin{tabular}{c|c|c|c|c|c|c|c|c|c}
        \multicolumn{2}{c|}{} & \multicolumn{2}{c|}{\textbf{Freeway}} & \multicolumn{2}{c|}{\textbf{Breakout}} & \multicolumn{2}{c|}{\textbf{Enduro}} & \multicolumn{2}{c}{\textbf{PacMan}} \\ \hline\hline
        \multirow{2}{*}{\textbf{Benchmark} ($\pi_{\text{ext}}$)} &  & Med & Good & Med & Good & Med & Good & Med & Good \\
        \cline{2-10}
         & $\hat J^{\pi_{\text{ext}}}_{-}$ & $15$ & $24$ & $61$ & $83$ & $351$ & $487$ & $437$ & $630$ \\ \hline\hline
        \multirow{2}{*}{\textbf{EVEREST (ours)}} & Return $\uparrow$ & $\mathbf{33 \pm 0}$ & $\mathbf{34 \pm 0}$ & $\mathbf{257 \pm 60}$ & $\mathbf{286 \pm 54}$ & $\mathbf{1285 \pm 154}$ & $\mathbf{1370 \pm 131}$ & $\mathbf{1993 \pm 301}$ & $\mathbf{1944 \pm 321}$\\
        \cline{2-10}
         & Regret $\downarrow$ & $\mathbf{18 \pm 11}$ & $\mathbf{86 \pm 4}$ & $\mathbf{965 \pm 36}$ & $\mathbf{497 \pm 103}$ & $\mathbf{19 \pm 2}$ & $\mathbf{48 \pm 1}$ & $\mathbf{239 \pm 21}$ & $\mathbf{92 \pm 7}$ \\
         \hline
        \multirow{2}{*}{\textbf{Oblivious}} & Return $\uparrow$ & \multicolumn{2}{c|}{$\mathbf{34 \pm 0}$} & \multicolumn{2}{c|}{$\mathbf{253 \pm 57}$} & \multicolumn{2}{c|}{$793 \pm 105$} & \multicolumn{2}{c}{$\mathbf{2002 \pm 220}$} \\
        \cline{2-10}
         & Regret $\downarrow$ & $721 \pm 51$ & $900 \pm 47$ & $3819 \pm 123$ & $4265 \pm 199$ & $507 \pm 17$ & $646 \pm 15$ & $891 \pm 11$ & $1247 \pm 41$ \\
    \end{tabular}
    }
    \caption{\textbf{External stability in Atari:} EVEREST is capable of reaching equal or higher performance, when compared to an oblivious learner, whilst dramatically reducing external instability (regret). Specifically, EVEREST attains $70\%$ higher returns on Enduro, when compared to the baseline.}
    \label{tab: process safety Atari}
\end{table*}

We compare four methods, as follows. \textbf{(1) Baseline:} We report the performance of the baseline TD3 algorithm, which updates the target network at each step in a Polyak averaging procedure. \textbf{(2) Max without reevaluation:} This na\"ive update scheme records the measured performance of the current target-network (without performing reevaluation). The update occurs if $\hat J_{\pi_\theta} \geq \hat J_{\pi_{-}}$. While this scheme is sample efficient, it results in a biased estimator of the performance. In turn, it is expected to suffer from high variance. \textbf{(3) Max with reevaluation:} A similar update scheme to `Max without reevaluation' that only compares the sampled mean. However, to ensure an unbiased process, the target network performance is reevaluated each time. \textbf{(4) EVEREST:} Our proposed method that periodically evaluates both the target and online policy. By performing a statistical test, taking into consideration both the means and standard deviations, we can ensure, with high probability, that the online policy improves upon the target. As such, only if the statistical test passes, the target network is updated.

\paragraph{Evaluation Metrics.} To evaluate internal stability, in addition to the standard median and mean performance, we consider reliability metrics \cite{rl_reliability_metrics} such as (i) dispersion and (ii) risk.

Dispersion measures the width of a distribution. We measure the inter-quartile range (IQR), specifically, the difference between the 25th and 75th percentiles. As the goal of reliability is to isolate higher frequencies, \citet{rl_reliability_metrics} propose to de-trend the data ($y_t' = y_t - y_{t-1}$). The IQR is presented for 3 sequential phases of the learning process -- beginning, middle, and end.

As the IQR considers the center of the distribution (25th to 75th percentiles), we also present a complementary risk measure -- the conditional value at risk (CVaR). This risk measure considers the expected loss in the worst case scenarios, i.e., $\CVaR_\alpha (X) = \mathbb{E} [ X | X \leq \VaR_\alpha (X) ]$, where $\alpha \in (0, 1)$ and $\VaR_\alpha (X)$ is the $\alpha$ quantile of $X$. Similarly to IQR, the CVaR is measured over de-trended differences. We measure the CVaR over the entire training process.

Based on these metrics, algorithms that perform well are less likely to suddenly suffer from performance degradation (e.g., forgetting previously learned behaviors). These are presented in \Cref{fig: comparisons2} and \Cref{tab: process reliability}. As well, in \Cref{fig: partial internal stability}, we compare the oblivious learner (TD3 baseline) with EVEREST. In the supplementary material, we present the remaining learning curves, including those for the other flavors, which we have proposed.

\paragraph{Results.} Using the proposed evaluation metrics, we describe results for internal stability, as depicted in \Cref{fig: median,fig: iqr,fig: comparisons} and \Cref{tab: process reliability}.

\textit{Median and mean performance during training} (\Cref{fig: median} and \Cref{tab: process reliability}): While the baseline (TD3 with Polyak averaging) has better initial performance; overtime our proposed methods for internal stabilization surpass it, where the best performing (in terms of raw performance) is the statistically-sound EVEREST update rule. We also report the mean and standard deviation over the last 10 policy evaluations (\Cref{tab: process reliability}). This shows that, while all methods achieve similar final mean performance, the EVEREST update scheme dramatically reduces instability and increases reliability.

\textit{Inter Quartile Range (IQR) across time} (\Cref{fig: iqr}): All three target-network update schemes outperform the baseline. Therefore, replacing Polyak with EVEREST significantly improves the stability of the learning process. Moreover, EVEREST has the highest level of stability of our proposed methods, also illustrated by the individual learning curves in the appendix.

\textit{Lower Conditional Value at Risk (CVaR) on Differences} (\Cref{fig: comparisons}): Based on these results, all three methods significantly increase reliability compared with a baseline. This quantifies the line plots and emphasizes the minimization of performance drawdowns throughout training by minimizing the average worst case (25 percentile) drawdown. Even though all three schemes improve stability, no one method outperforms the others, in all individual tasks.

\subsection{External Stability}\label{sec: exp external}

In order to be externally stable, an agent must guarantee that it will improve in accordance with a given benchmark policy with a high probability. To obtain a benchmark policy, we train a standard agent and take snapshots midway through the training process (SAC or DQN for MuJoCo and Atari, respectively). 

Using an ensemble of value estimators, we evaluate the frozen benchmark policy. As a result, the mean and standard deviation of the samples are calculated over the ensemble, i.e., $\hat v^{\pi_{\text{ext}}} (\state) = \frac{1}{N} \sum_{i=1}^N \hat v^{\pi_{\text{ext}}}_i (\state)$ and ${\hat \sigma_{v^{\pi_{\text{ext}}}} (\state) = \sqrt{\frac{1}{N} \sum_{i=1}^N (\hat v^{\pi_{\text{ext}}}_i (\state) - \hat v^{\pi_{\text{ext}}} (\state))^2}}$.

To achieve external stability, EVEREST utilizes an admissible action set~$\bar{\mathcal{A}}$ (\Cref{eqn: admissible action set}), defined by a lower confidence bound on the $Q$-function and benchmark value. We use an online ensemble of $Q$-function estimators to estimate the lower confidence bound per action, i.e., $\hat Q_{-}^\pi (\state, \action) = \hat Q^\pi (\state, \action) - C \cdot \hat \sigma_{Q^\pi} (\state, \action)$, where $C$ is a predefined constant. Then, the admissible action set is constructed by evaluating every action. For continuous action spaces we approximate this set by sampling a fixed number of actions from the policy $\pi$ and constructing an admissible action set from them. Finally, the agent acts according to the admissible action set $\bar{\mathcal{A}}$. In this work, we consider the $\epsilon-$greedy scheme, where the agent randomly selects an action within $\bar{\mathcal{A}}$ with probability $\epsilon$, and with probability $1 - \epsilon$ acts greedily, i.e., $\argmax_{\action \in \bar{\mathcal{A}}} \hat Q^\pi (\state, \action)$.

The results and lower performance bounds are summarized in \Cref{tab: process safety mujoco,tab: process safety Atari}. We also present a sample of the plots in \Cref{fig: external stability miniplots}, whereas the rest are provided in the supplementary material. For analysis, we measure the return and regret. While the return is a periodic measurement of the policy performance, the regret analyzes the sub-optimality of the agent. Specifically, define the regret as the cumulative number of episodes (from the start of learning and up to time $t$) in which the agent has performed sub-optimally w.r.t. the benchmark\footnote{In the context of PAC-MDPs this is also known as the sample complexity \citep{kakade2003sample}. However, as we consider stability and reliability, regret is a more fitting term.}, i.e.,
\begin{equation*}
    \mathcal{R}_t (\pi) = \sum_{k=1}^t \indicator{J^{\pi_k} < J^{\pi_{\text{ext}}}} \,.
\end{equation*}

\begin{figure}[t]
    \centering
    \textbf{Evolution Of The Admissible Action Space}\\
    \includegraphics[width=0.55\linewidth]{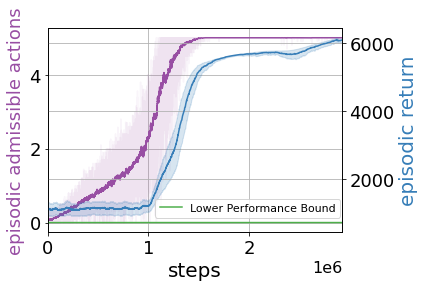}
    \caption{\textbf{External Stability:} Evolution of the admissible action space size, for EVEREST in Humanoid, over time. The light lines represent the various seeds, whereas the opaque line shows the smoothed average across seeds. At each episode throughout training, we report the average size of the admissible action space in addition to the obtained reward. The lower performance bound represents the safety threshold above which the agent should remain. As the learning process advances, the agent becomes confident, more actions are admissible and its performance improves.}
    \label{fig: action size evolution}
\end{figure}

\textit{Evolution of the admissible action space:} \Cref{fig: action size evolution} presents the evolution of the admissible action space over time, alongside a plot of the learner's performance. Each point represents the average over an entire episode. As EVEREST maintains a conservative action space, i.e., the set of actions that are improving w.h.p., this set is initially empty. Over time, the agent improves and becomes more confident, increasing the number of admissible actions. The reliability effects are seen throughout the entire learning process. We observe that even when the learner outperforms the benchmark, it may still utilize it (rarely) to ensure optimal performance. We attribute this to the exploratory nature of the underlying learner, as it is continually exploring, it may reach novel states or states it has forgotten.

\textit{Regret:} As the admissible action set only contains actions that are w.h.p. at least as good as the benchmark, EVEREST exhibits lower regret across all tested scenarios. However, as EVEREST follows a probabilistic mechanism, it does not ensure zero violations. Hence, even though it attains a much lower regret, this regret is strictly positive.

\textit{Process performance:} In addition to analyzing the regret, it is important to consider the actual underlying performance of the agent. An overly pessimistic agent may continually pass control to the benchmark and never become confident enough to take control and improve. What we observe is the opposite. Not only does EVEREST slowly take control and outperform the benchmark, but in all tasks, it exhibits performance at least as good as the oblivious learner. In addition, in some domains (Enduro and Ant) EVEREST exhibits superior performance compared to the oblivious learner ($70\%$ and $40\%$ increase in performance, respectively). In \Cref{fig: external stability miniplots} we present example behavior from two environments, showing how EVEREST minimizes the regret and ensures a high probability of lower bound performance.

\begin{figure}
    \centering
    \hspace{0.5cm}\textbf{Return}\hspace{3.2cm}\textbf{Regret}\vspace{0.1cm} \\
    \includegraphics[width=0.49\linewidth]{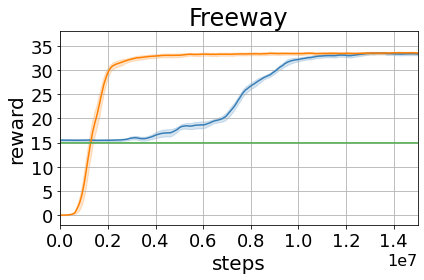}
    \includegraphics[width=0.49\linewidth]{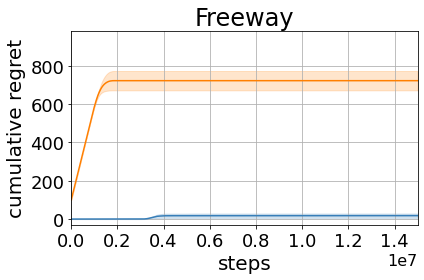}
    \includegraphics[width=0.49\linewidth]{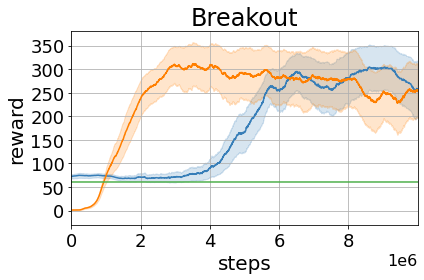}
    \includegraphics[width=0.49\linewidth]{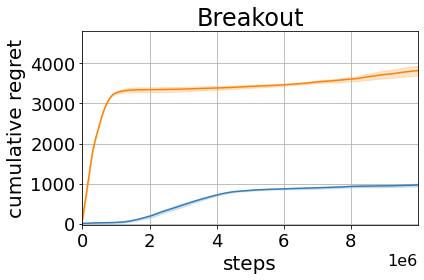}\\
    \includegraphics[width=\linewidth,trim={0.1cm 0 0 0},clip]{figures_safety/legend}
    \caption{\textbf{External stability:} we compare EVEREST with an oblivious learner (here, DQN). EVEREST minimizes the benchmark regret, ensuring, w.h.p., performance at least at the lower performance bound.}
    \label{fig: external stability miniplots}
\end{figure}

\section{Related Work}

\paragraph{Monotonous improvement of learning algorithms:} montonticity is a desirable property for many machine learning algorithms. Recently, \citet{viering2019open} studied this question in the setting of empirical risk minimization. They questioned whether adding a learner would exhibit monotone improvement when presented with an additional sample. Surprisingly, while this holds in certain problems, they also presented several tasks in which it does not. \citet{bousquet2022monotone} then presented a method, for multi-class classification, in which any learner can be monotonised -- e.g., converted into a monotone learner.


\paragraph{Imitation Learning:} An intuitive way of learning an externally stable policy is using imitation learning techniques. In imitation learning, the goal of the learner is to obtain a policy that is as similar to the benchmark. Specifically, active learning methods, such as DAgger \citep{ross2011reduction}, can ensure a high probability of attaining a lower bound on the benchmark's performance. However, as they perform imitation learning, they will not learn to further improve upon the learner.

\paragraph{Safety in Deep RL:} The pitfall of existing exploration schemes is their ignorance of catastrophic outcomes throughout learning, a critical issue when considering training RL agents in the real world \cite{leike2017ai}. While theoretical works have tackled this problem, they often require information that is impossible to obtain -- e.g., a precise model. Hence, works have focused on myopic safety with a local model \citep{dalal2018safe} or constrained optimization schemes that aren't assured to behave well throughout training but eventually converge to a feasible policy \citep{achiam2017constrained,tessler2018reward}.

\paragraph{Safe Policy Improvement:} \cite{thomas2015high} proposed a notion of safety relative to a given benchmark policy, yet this was only enforced on the final produced policy (not throughout training). Later on, \cite{laroche2017safe} presented the notion of safe policy improvement with benchmark boostrapping, here, they consider safe policy improvement in the batch RL setting (offline RL). In simple terms, they switch between bootstrapping on the learner and the benchmark based on whether an $(s, a)$ pair is observed sufficiently often.

\section{Conclusions}

Reinforcement learning algorithms are notoriously unstable and unreliable \citep{henderson2018deep,rlblogpost}. As recent advances have focused on reproducibility, many of these concerns have been alleviated \citep{schulman2017proximal,haarnoja2018soft,fujimoto2018addressing,badia2020agent57}; however, since they emphasize the learning trend, they present smoothed learning curves, creating a false impression of stability.

In this work we present two stability issues for RL agents, which we termed internal and external stability. While internal stability refers to the agent and its historical performance with respect to itself, external stability, akin to safety, considers stability with respect to an external policy.

We presented a simple to implement, yet theoretically justified, method which we call EVEREST. We performed extensive evaluations on both Atari and MuJoCo. Using reliability metrics proposed by \cite{rl_reliability_metrics}, we measured internal stability and observed that EVEREST is indeed more reliable and stable across all domains; improving converged performance across most domains. External stability was measured using a strict sub-optimality count. Similarly, we observed a dramatic improvement in stability (reduction in regret) and, surprisingly, an improvement in final converged performance across several tasks.

\bibliography{bibliography}

\clearpage

\onecolumn
\appendix

\section{Improvement of mixture policy -- finite horizon setting}
\begin{theorem}\label{thm: finite horizon}
    Let $\brk[c]*{\pi_{i}}_{i=1}^M$ and define $\bar{\pi}$ such that for all ${s \in \S}$, $\bar{\pi}_h(s) \in \arg\max_{i \in [M]} v_h^{\pi_{i}}(s)$. Then, 
    \begin{align*}
    v_h^{\bar{\pi}_h}(s) \geq \max_{i \in [M]} v_h^{\pi_{i}}(s), \forall s \in \S, h \in [H].
    \end{align*}
\end{theorem}
\begin{proof}
    We begin by defining an auxiliary policy which at state $s$ and time step $h$ chooses the best policy from the set $\brk[c]*{\pi_i}_{i=1}^M$ and stays fixed. That is,
    \begin{align*}
        \mu_h(s) = \arg\max_{i \in [M]} v_h^{\pi_i}(s).
    \end{align*}
    Given multiple maximizers, the maximizer is chosen to be the same as in $\bar{\pi}$. 
    Notice that it is not necessarily true that $v_h^{\bar{\pi}}(s) = v_h^{\mu_h}(s)$ for $h \in [H]$. By definition, we have that $v_h^{\mu_h}(s) \geq v_h^{\pi_i}(s)$ for all $i \in [M], s \in \S$. Therefore, it remains to show that $v_h^{\bar{\pi}_h}(s) \geq v_h^{\mu_h}(s)$ for all $h \in [H]$. We prove this by induction on $h \in [H]$.
    
    The base case follows trivially for $h = H - 1$. Next, assume that $v^{\bar{\pi}}_{h+1}(s) \geq v^{\mu_{h+1}}_{h+1}(s)$ holds for all $s \in \S$. We have that
    \begin{align*}
        v_h^{\bar{\pi}}(s) 
        &= 
        r_h(s, \bar{\pi}_h(s)) + \sum_{s' \in \S}P_h(s'|s,\bar{\pi}_h(s))v_{h+1}^{\bar{\pi}}(s') \\
        &\geq
        r_h(s, \bar{\pi}_h(s)) + \sum_{s' \in \S}P_h(s'|s,\bar{\pi}_h(s))v_{h+1}^{\mu_{h+1}}(s') \\
        &=
        r_h(s, \mu_h(s)) + \sum_{s' \in \S}P_h(s'|s,\mu_h(s))v_{h+1}^{\mu_{h+1}}(s') \\
        &\geq
        r_h(s, \mu_h(s)) + \sum_{s' \in \S}P_h(s'|s,\mu_h(s))v_{h+1}^{\mu_{h}}(s') \\
        &=
        v_h^{\mu_h}(s).
    \end{align*}
    where the first relation follows by the induction step, the second relation follows by definition of $\mu_h$, and $\bar{\pi}_h$, which are equal at time step $h$, and the third relation is since, by definition, $v_{h+1}^{\mu_{h+1}}(s)=\max_{i\in\brs{M}}v_{h+1}^{\pi_i}(s)\ge v_{h+1}^{\mu_{h}}(s)$ for all $s\in\S$ (as $\mu_h\in\brc*{\pi_i}_{i=1}^M$). This completes the proof.
\end{proof}

\section{Improvement of mixture policy -- discounted setting}

\begin{theorem}
    Let $\brk[c]*{\pi_{i}}_{i=1}^M$ and define $\bar{\pi}$ such that for all ${s \in \S}$, $\bar{\pi}(s) \in \arg\max_{i \in [M]} v^{\pi_{i}}(s)$. Then,
    \begin{align*}
    v^{\bar{\pi}}(s) \geq \max_{i \in [M]} v^{\pi_{i}}(s), \forall s \in \S.
    \end{align*}
\end{theorem}
\begin{proof}
    Denote $i(s)\in \argmax_{i\in\brs*{M}} v^{\pi}_i(s)$, and given multiple maximizers, assume that the maximizer is chosen to be the same as in $\bar{\pi}$. Then, our goal is to prove that $v^{\bar{\pi}}(s) \geq \max_{i\in\brs*{M}} v^{\pi_{i}}(s)\triangleq v^{\pi_{i(s)}}(s)$ for any $s\in\S$. To do so, we define the non-stationary policy $\bar{\pi}^{k}_h$ such that $\bar{\pi}^k_h(s)=\bar{\pi}(s)$ for any $h\le k$ and $\bar{\pi}^k_h(s)=\pi_{i(s_k)}(s)$ for any $h>k$. In particular, by definition, $\bar{\pi}^0_h(s)=\pi_{i(s)}$, so it clearly holds that $v^{\bar{\pi}^0}(s) \geq v^{\pi_{i(s)}}(s)$ for all $s\in\S$. 
    
    We next show that the value of $\bar{\pi}^k$ is nondecreasing, namely, $v^{\bar{\pi}^k_h}(s)\geq v^{\bar{\pi}^{k-1}_h}(s)$ for all $s\in\S$ and $k>0$. To do so, we use a coupling argument; namely, notice that up to time $k-1$, the policies $\bar{\pi}^{k-1}$ and $\bar{\pi}^{k}$ are identical. Therefore, we can run both policies using the same internal randomization of the environment and policy, such that given any initial state $s$, both policies will visit the same trajectory up to state $s_k$ (included). Formally, letting $s_h$ and $s'_h$ be the states visited by $\bar{\pi}^{k-1}$ and $\bar{\pi}^{k}$, respectively, we choose the randomization such that given an initial state $s$, it holds that $s_h=s'_h$ for all $h\leq k$. 
    In particular, with a slight abuse of notations, we treat $\pi(s)$ as the random action taken by the policy $\pi$ at state $s$ and couple the trajectories such that $\bar{\pi}^{k-1}_h(s_h)=\bar{\pi}^{k}_h(s_h)$ for all $h\leq k-1$. 
    Then, we can write
    \begin{align*}
        v^{\bar{\pi}^{k-1} }(s)
        &= 
        \E\brs*{\sum_{h=0}^\infty \gamma^hr\br*{s_h,\bar{\pi}^{k-1}_h(s_h)}\vert s_0=s}\\
        &\overset{(1)}{=}
        \E\brs*{\sum_{h=0}^{k-1} \gamma^hr\br*{s_h,\bar{\pi}^{k-1}_h(s_h)}\vert s_0=s} + \E\brs*{\sum_{h=k}^\infty \gamma^hr\br*{s_h,\pi_{i(s_{k-1})}(s_h)}\vert s_0=s} \\
        &\overset{(2)}{=}
        \E\brs*{\sum_{h=0}^{k-1} \gamma^hr\br*{s_h,\bar{\pi}^{k-1}_h(s_h)}\vert s_0=s} + \gamma^k\E\brs*{v^{\pi_{i(s_{k-1})}}(s_k)\vert s_0=s} \\
        &\overset{(3)}{\leq}
        \E\brs*{\sum_{h=0}^{k-1} \gamma^hr\br*{s_h,\bar{\pi}^{k}_h(s_h)}\vert s_0=s} + \gamma^k\E\brs*{v^{\pi_{i(s_{k})}}(s_k)\vert s_0=s} \\
        &\overset{(4)}{=}
        \E\brs*{\sum_{h=0}^{k-1} \gamma^hr\br*{s'_h,\bar{\pi}^{k}_h(s'_h)}\vert s_0=s} + \gamma^k\E\brs*{v^{\pi_{i(s'_{k})}}(s'_k)\vert s_0=s} \\
        &=
        \E\brs*{\sum_{h=0}^{k-1} \gamma^hr\br*{s'_h,\bar{\pi}^{k}_h(s'_h)}\vert s_0=s} + \E\brs*{\sum_{h=k}^\infty \gamma^hr\br*{s'_h,\pi_{i(s'_{k})}(s'_h)}\vert s_0=s} \\
        &=
        v^{\bar{\pi}^k }(s)\enspace.
    \end{align*}
    $(1)$ is by the definition of $\bar{\pi}^{k-1}_h$ and $(2)$ is by the definition of the value of $\pi_{i(s_{k-1})}$. $(3)$ is since $\bar{\pi}^{k-1}_h$ and $\bar{\pi}^{k}_h$ are the same for $h<k$ and since, by definition, $v^{\pi_{i(s_{k})}}(s_k) = \max_i v^{\pi_{i}}(s_k) \geq v^{\pi_{i(s_{k-1})}}(s_k)$. Finally, $(4)$ is by the coupling argument -- both policies pass through the same trajectory up to $s_k$. Therefore, combined with the fact that $v^{\bar{\pi}^0}(s) \geq v^{\pi_{i(s)}}(s)$ for all $s\in\S$, the monotonicity implies that $v^{\bar{\pi}^{k}}(s) \geq v^{\pi_{i(s)}}(s)$ for any $k\in\mathbb{N}$ and $s\in\S$.

    To finalize the proof, it remains to show that $\lim_{k\to\infty} v^{\bar{\pi}^k }(s) = v^{\bar{\pi}}(s)$; then, as $v^{\bar{\pi}^{k}}(s) \geq v^{\pi_{i(s)}}(s)$ for any $k$, it would also holds for the limit. As in the previous derivation, we assume that all $\bar{\pi}^k$ are coupled to $\bar{\pi}$, i.e., follow the same trajectory as $\bar{\pi}$ until $s_k$ and only diverge from the trajectory after the policy changes. This choice of the probability space naturally implies that $\lim_{k\to\infty}r\br*{s_h,\bar{\pi}^k_h(s_h)} = r\br*{s_h,\bar{\pi}(s_h)}$ for any $h\in\mathbb{N}$. Then, we have that
    \begin{align*}
        v^{\bar{\pi}}(s)
        &=
        \E\brs*{\sum_{h=0}^\infty \gamma^hr\br*{s_h,\bar{\pi}(s_h)}\vert s_0=s} \\
        &=
        \E\brs*{\sum_{h=0}^\infty \lim_{k\to\infty}\gamma^hr\br*{s_h,\bar{\pi}^k_h(s_h)}\vert s_0=s} \\
        &\overset{(1)}{=}
        \E\brs*{\lim_{k\to\infty}\sum_{h=0}^\infty \gamma^hr\br*{s_h,\bar{\pi}^k_h(s_h)}\vert s_0=s} \\
        &\overset{(2)}{=} 
        \lim_{k\to\infty}\E\brs*{\sum_{h=0}^\infty \gamma^hr\br*{s_h,\bar{\pi}^k_h(s_h)}\vert s_0=s} \\
        & = 
        \lim_{k\to\infty} v^{\bar{\pi}^k }(s)\enspace.
    \end{align*}
    Both $(1)$ and $(2)$ hold by the bounded convergence theorem. In particular, for $R_{\max}=\max_{(s,a)\in\S\times\aset} \abs{r(s,a)}$, we have that $\abs{\gamma^hr\br*{s_h,\bar{\pi}^k_h(s_h)}}\le \gamma^h R_{\max}$ and $\sum_{h=0}^\infty \gamma^hR_{\max}=\frac{R_{\max}}{1-\gamma}<\infty$, so the limit and sum are interchangeable and $(1)$ holds. Similarly, we can bound $\sum_{h=0}^\infty \abs{\gamma^hr\br*{s_h,\bar{\pi}_h(s_h)}} \leq \frac{R_{\max}}{1-\gamma}<\infty$, which make the expectation and limit interchangeable.
    

\end{proof}

\clearpage
\section{Experiments}

All experiments were performed on a set of machines with NVIDIA GTX 1080 GPUs and 12 core Intel i7 CPUs.

As we focus on improving stability, our goal is to avoid changing hyper parameters unless required. Hence, for all methods we follow the standard hyper parameters reported for TD3 \citep{fujimoto2018addressing}, SAC \citep{haarnoja2018soft} and Bootstrapped DQN \citep{osband2016deep}.

For internal stability we evaluate and swap (based on the t-test) every 10,000 steps. Each policy is evaluated for 10 episodes and the swap occurs if the probability the policy improved is over $90\%$. Typically, papers report the performance of the online network. We observed that the target network attains similar performance but with much higher stability. Hence, we compare to the reliability of the target network, updated using Polyak-averaging, and not that of the online policy.

For external stability we maintain an ensemble of 5 $Q-$estimators. At each state, the admissible action set is determined and the agent plays $\epsilon-$greedy on this set. For MuJoCo domains we sample $5$ actions from the policy and construct $\bar{\mathcal{A}}$ for these actions.

To obtain the benchmark agent, we train an oblivious learner and take two snapshots midway -- the lower performing we deem `medium' and the better as `good'. We ensure that both are not perfect, to enable EVEREST to further learn and improve over the benchmark (as our goal is not imitation learning). The lower confidence estimate for $\pi_{\text{ext}}$ is obtained by training an ensemble of value-function estimates. While training $\pi$ we keep $v_i^{\pi_{\text{ext}}}$ static.

\subsection{Aggregated CVaR}

\begin{figure}[H]
    \centering
    \includegraphics[width=0.2\linewidth]{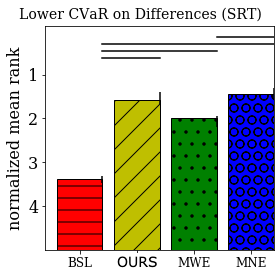}
    \caption{Aggregated (across domains) internal stability CVaR metrics. We compare BSL (Oblivious baseline learner, TD3), Ours (EVEREST), MWE (Max with reevaluation of the target network) and MNE (Max without reevaluation). All three flavors outperform the oblivious learner, however EVEREST and MNE perform similarly when aggregated across domains.}
    \label{fig: srt aggregated}
\end{figure}

In \Cref{fig: srt aggregated} we present the aggregated values for the CVaR on the drawdown. As seen in the per-environment graphs, EVEREST performs similarly to Max without reevaluation (when considering performance across multiple distinct environments). However, EVEREST, and the other two proposed flavors drastically outperform the oblivious learner.

\subsection{Alternative Approaches}

\begin{figure}[H]
    \centering
    \includegraphics[width=0.4\linewidth]{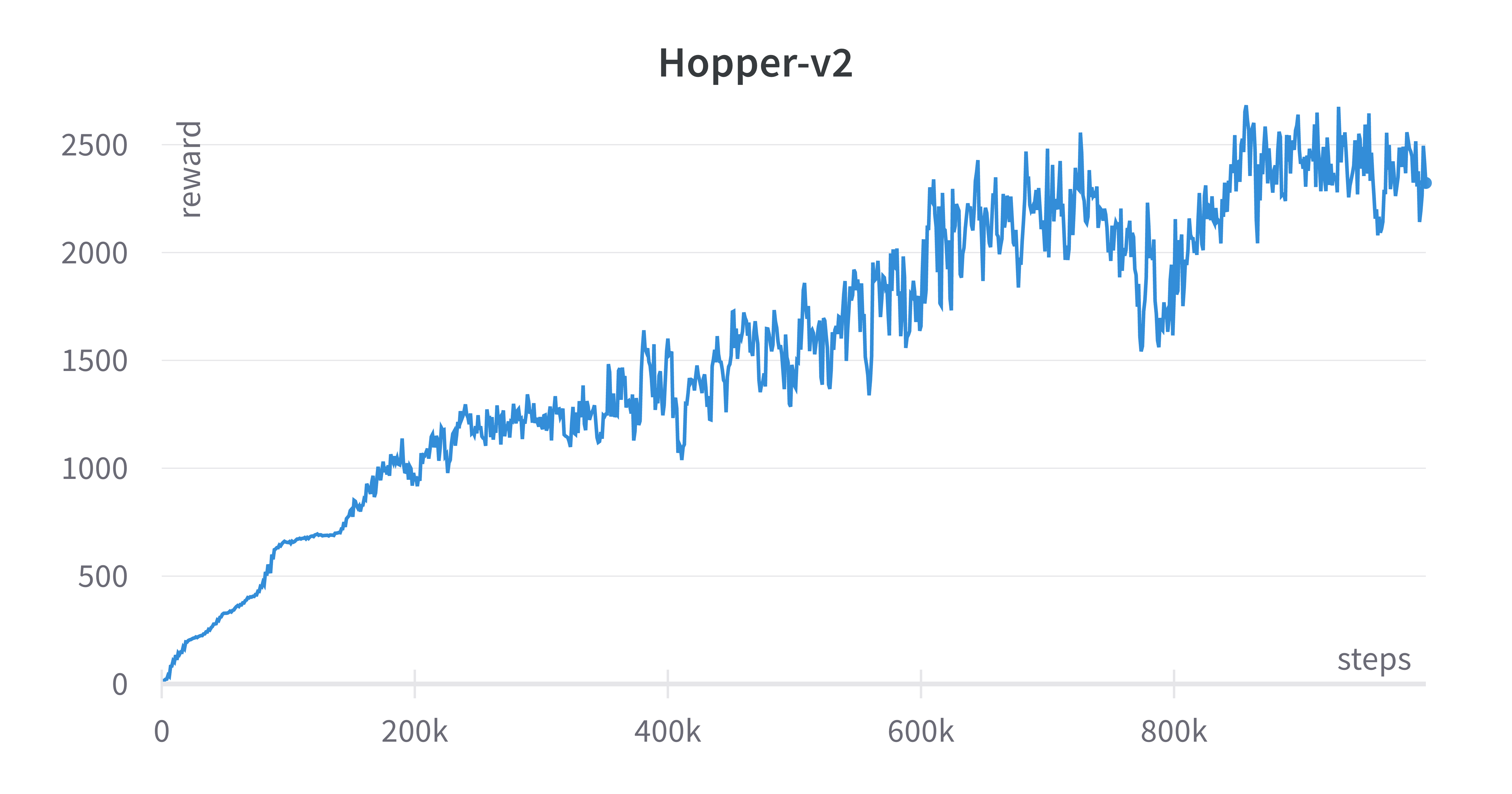}\hfill\includegraphics[width=0.4\linewidth]{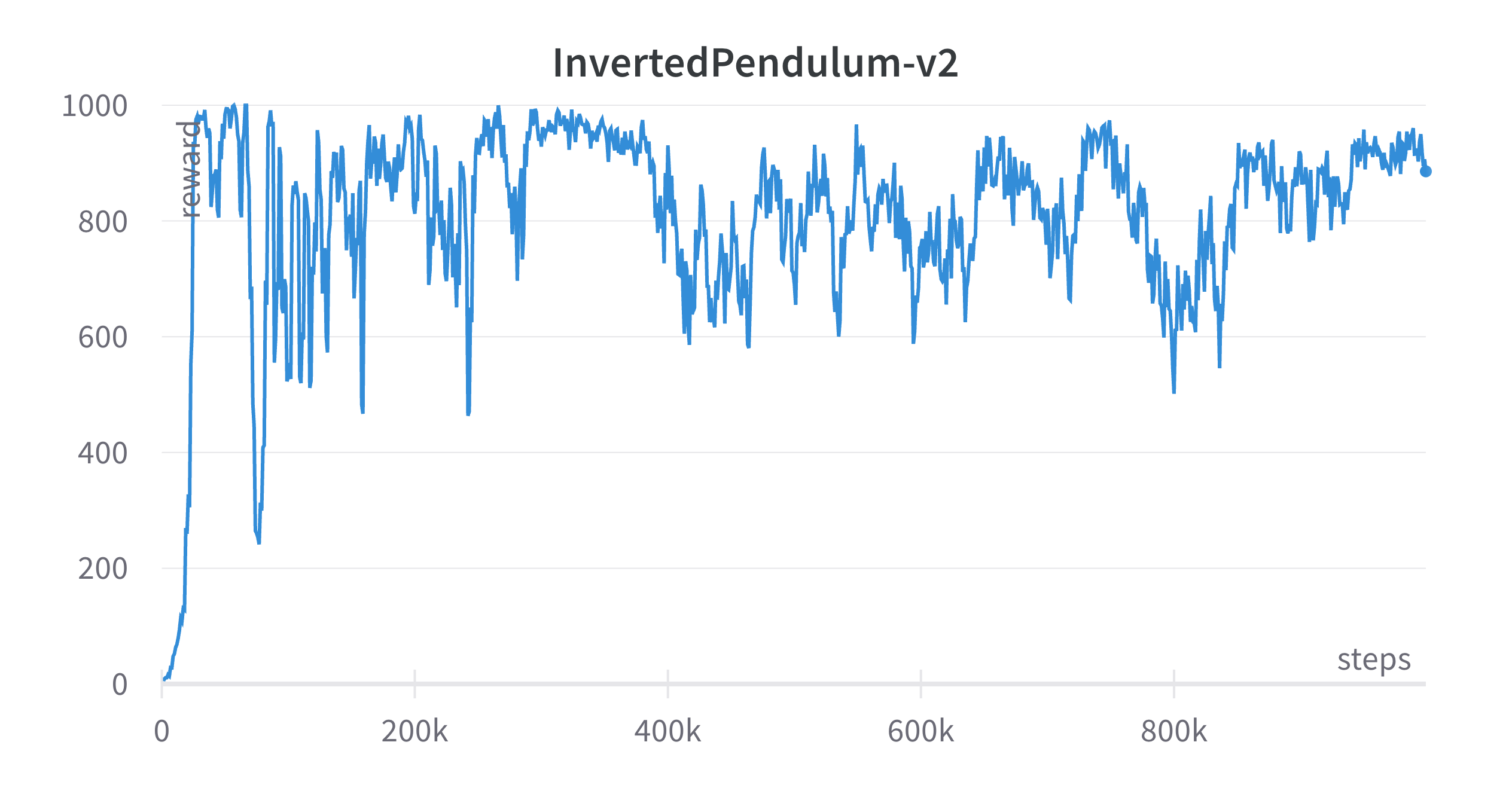}\\
    \includegraphics[width=0.4\linewidth]{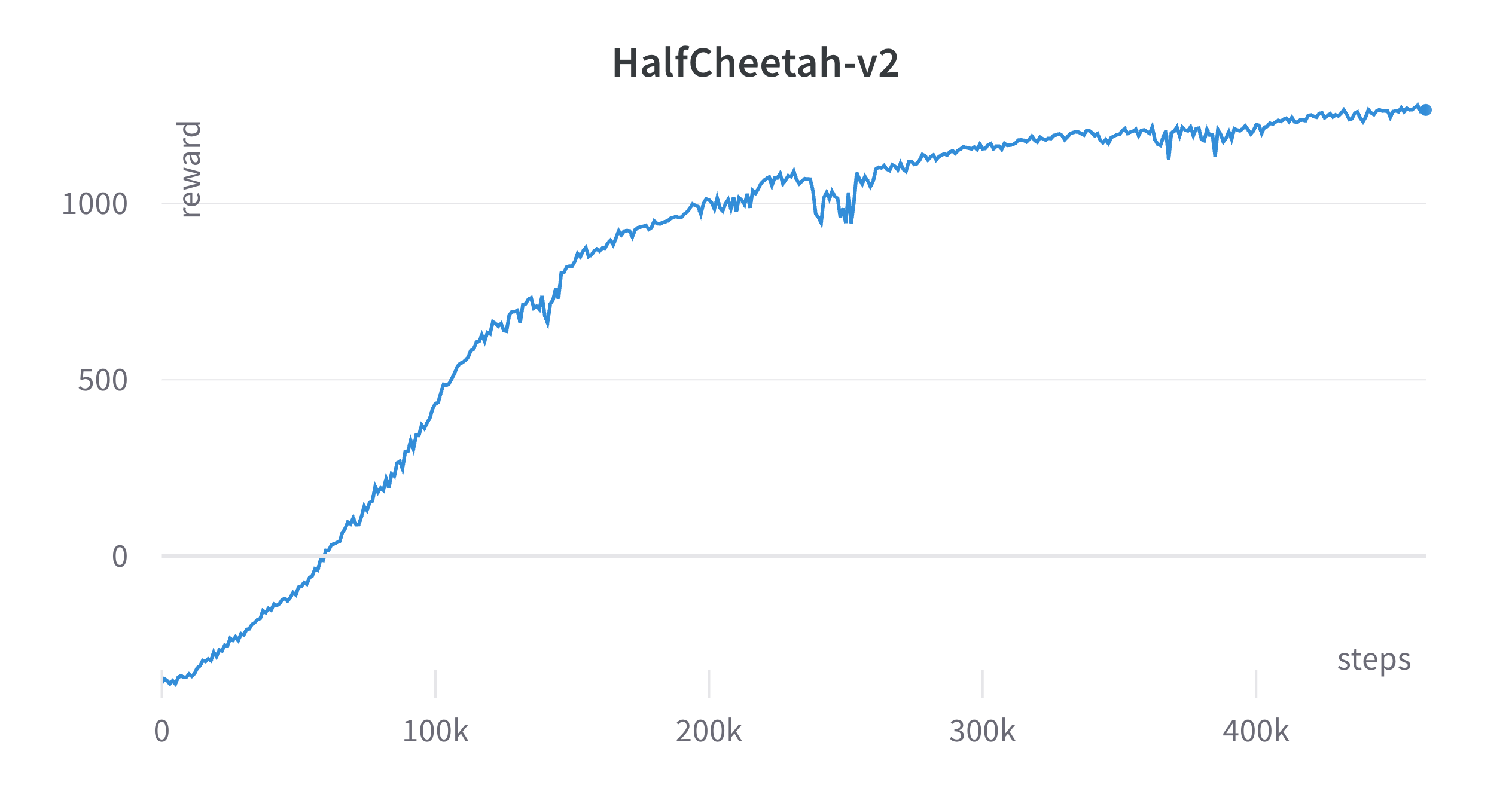}\hfill\includegraphics[width=0.4\linewidth]{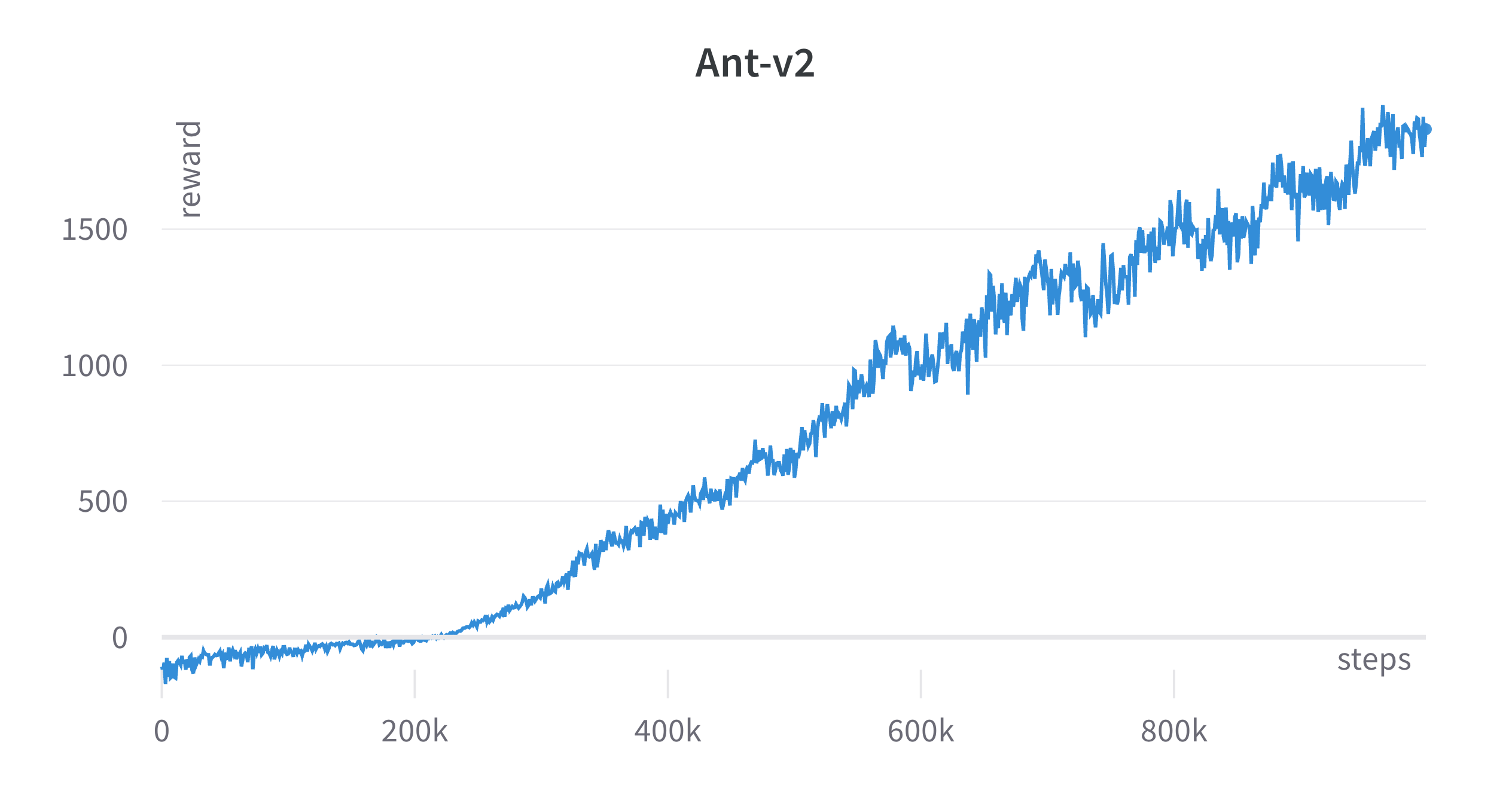}\\
    \includegraphics[width=0.4\linewidth]{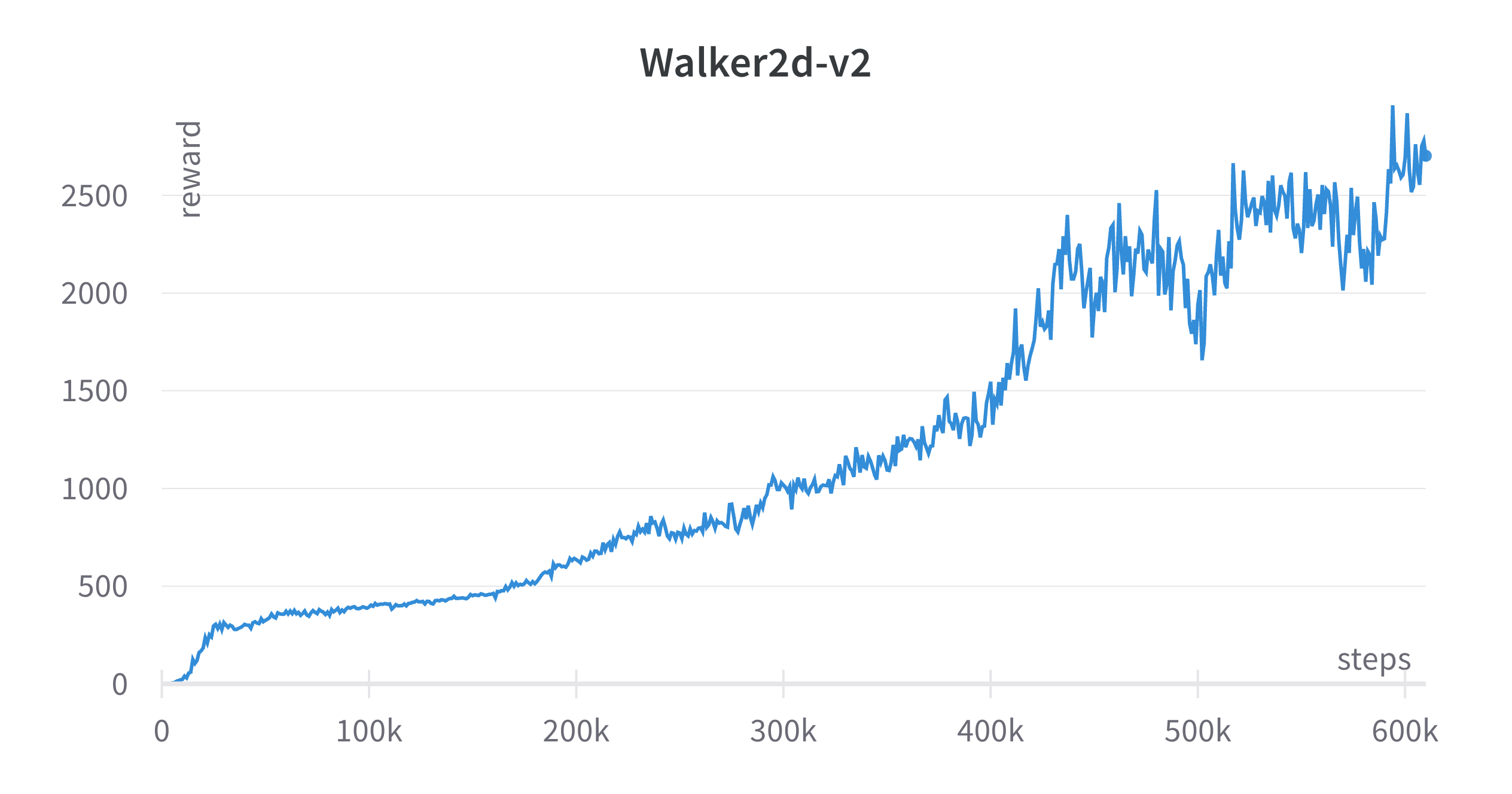}
    \caption{\textbf{PPO is internally unstable.} We present the \textit{non-smoothed} learning curves for PPO in the various environments. In each figure, a line represents a single seed. As can be seen, despite the trust region policy updates, the learning process is unstable.}
    \label{fig: bad ppo}
\end{figure}

\paragraph{Internal Stabilization:} Trust region schemes, such as PPO \citep{schulman2017proximal}, attempt to stabilize learning by limiting how much the policy may deviate. However, as they don't directly stabilize the value but rather attempt to limit the deviation of the policy. it is not ensured to be internally stable -- i.e., monotonously improving w.h.p. We present these results in \Cref{fig: bad ppo}. They show that in addition to PPO converging to lower final performance (compared to the baseline TD3 algorithm) it's stability varies between domains. While certain domains are simpler and stabler, e.g., HalfCheetah. In others we observe large swings and instability, for instance in the InvertedPendulum task. We can thus conclude that while PPO does attempt to minimize the policy deviation, \textit{it is not stable}.

\paragraph{External Stabilization:} We considered training the learners from scratch. One may wonder -- given access to a benchmark policy, why not initialize the learner using imitation learning? \textbf{Does a warm-start scheme improve external stability?}

\begin{figure}[H]
    \centering
    \includegraphics[width=0.35\linewidth]{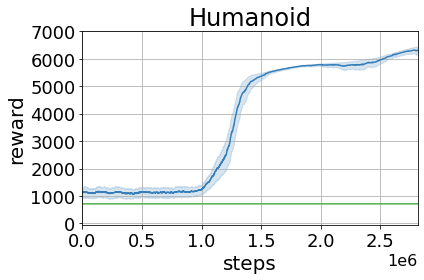}
    \caption{\textbf{EVEREST overcomes warmstart instability.} We initialize the agent using the benchmark policy. The learner then continues to learn by selecting actions as per EVEREST.}
    \label{fig: humanoid warmstart}
\end{figure}

\begin{figure}[H]
    \centering
    \includegraphics[width=0.5\linewidth]{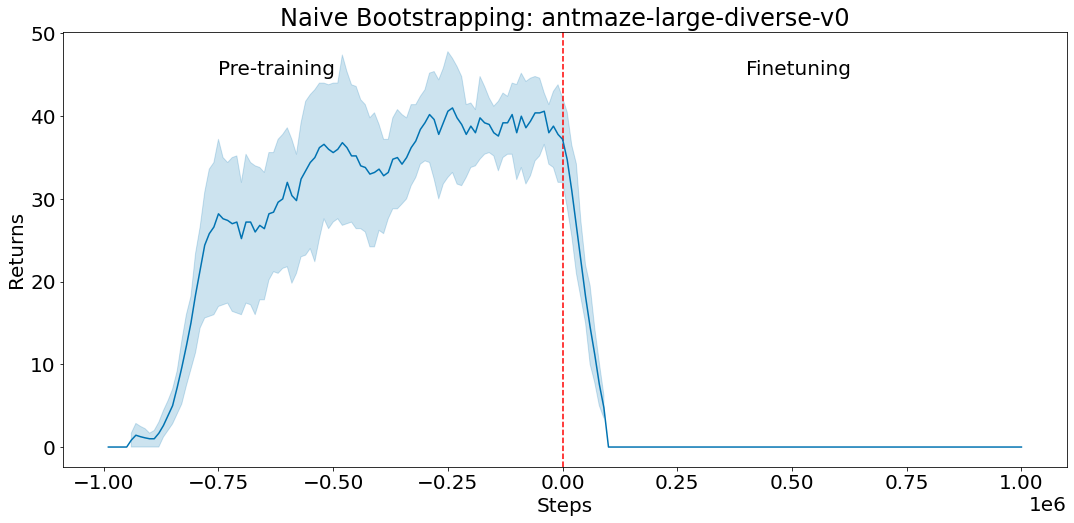}
    \caption{\textbf{The dangers of naıvely initializing the policy. \textit{Credit -- \citet{uchendu2022jump}.}} A policy is pre-trained on offline data to a medium level of performance. Negative steps correspond to this pre-training. Then, they use the policy to initialize actor-critic finetuning (positive steps starting from step 0.0) using this pretrained policy as the initial actor, and initializing the critic randomly. Actor performance immediately drops and does not recover, as the untrained critic provides an exceedingly poor learning signal, causing the good initial policy to be forgotten.}
    \label{fig: bad pre-training}
\end{figure}

We evaluated this on the Humanoid domain and present the results of EVEREST in \Cref{fig: humanoid warmstart}. An warm-started agent that is trained with EVEREST exhibits similar results to learning from scratch. On the other hand, we observe surprising results when warm-starting the oblivious learner. We observed catastrophic failures, in line with observations from prior work. \citet{uchendu2022jump} show that such a warm start scheme often fails catastrophically. An example is presented in \Cref{fig: bad pre-training}, taken from their paper.

We conclude that warmstarting does not overcome external stability and by using EVEREST we can ensure a minimal regret.

\subsection{Additional Results}

Below, we present the full training curves of all experiments. We begin with those for internal stability and continue with the external stability figures.

As can be seen, EVEREST improves stability across all experiments. This is apparent both quantitatively (as seen in the main paper) and qualitatively (as is observed visually).

\begin{figure*}
\centering
\begin{tabular}{>{\centering\arraybackslash}m{.22\linewidth} >{\centering\arraybackslash}m{.22\linewidth} >{\centering\arraybackslash}m{.22\linewidth} >{\centering\arraybackslash}m{.22\linewidth}}
 ~~EVEREST & ~~Max w/ reevaluation & ~~Max w/o reevaluation & ~~Baseline \\
 \includegraphics[width=\linewidth]{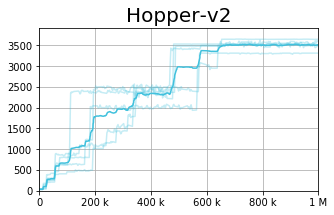} & 
 \includegraphics[width=\linewidth]{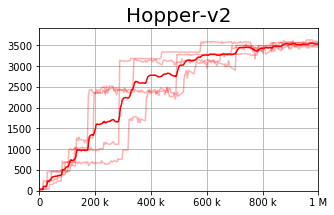} & \includegraphics[width=\linewidth]{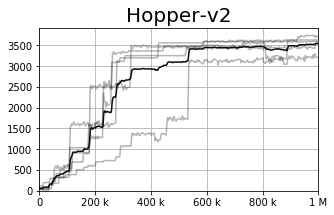} &
 \includegraphics[width=\linewidth]{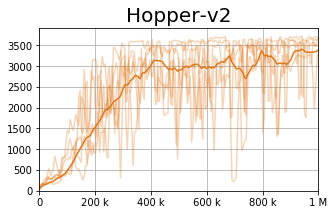} \\
 \includegraphics[width=\linewidth]{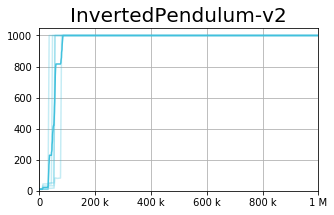} & \includegraphics[width=\linewidth]{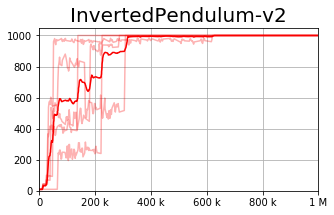} &
 \includegraphics[width=\linewidth]{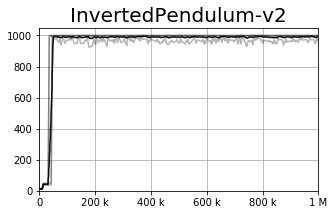} & 
 \includegraphics[width=\linewidth]{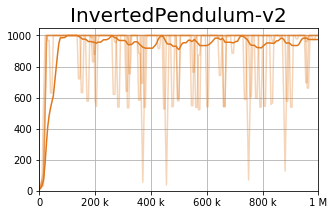}\\
 \includegraphics[width=\linewidth]{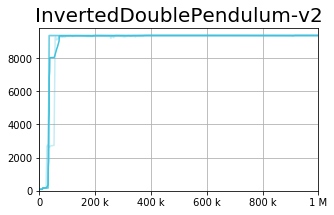} & 
 \includegraphics[width=\linewidth]{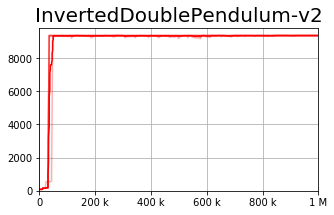} &
 \includegraphics[width=\linewidth]{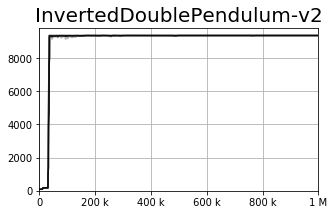} &
 \includegraphics[width=\linewidth]{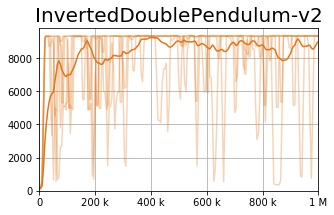} \\
 \includegraphics[width=\linewidth]{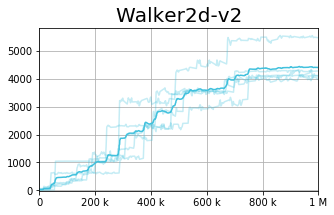} &
 \includegraphics[width=\linewidth]{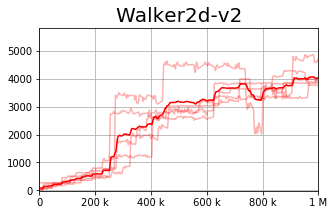} &
 \includegraphics[width=\linewidth]{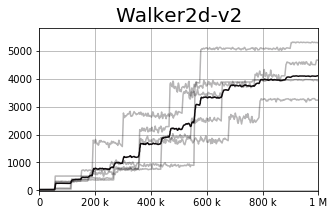} &
 \includegraphics[width=\linewidth]{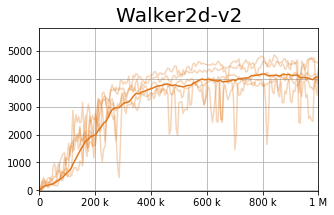} \\
 \includegraphics[width=\linewidth]{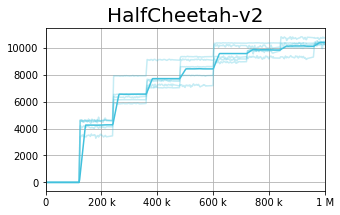} &
 \includegraphics[width=\linewidth]{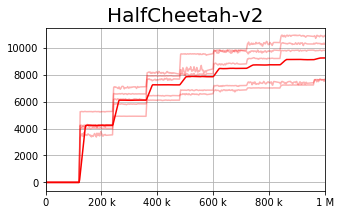} &
 \includegraphics[width=\linewidth]{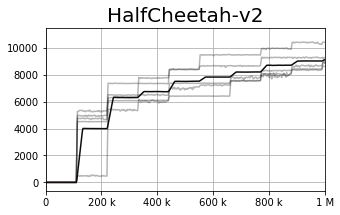} &
 \includegraphics[width=\linewidth]{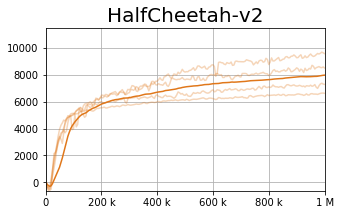}  \\
 \includegraphics[width=\linewidth]{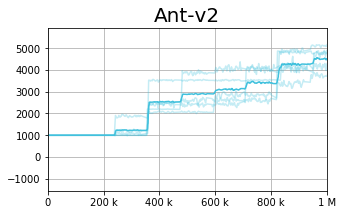} &
 \includegraphics[width=\linewidth]{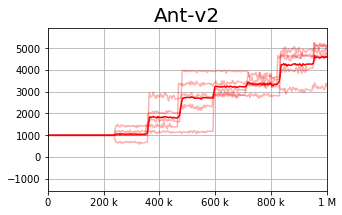} &
 \includegraphics[width=\linewidth]{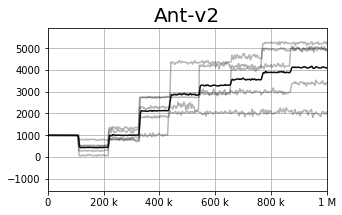} &
 \includegraphics[width=\linewidth]{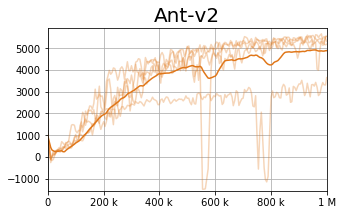}
\end{tabular}
\caption{\textbf{Internal stability:} x-axis denotes the total number of environment interactions. All 3 flavors improve stability across all environments. Comparing between the 3, we observe that EVEREST exhibits higher stability (see Walker2d and InvertedPendulum).}
\label{fig: results}
\end{figure*}

\begin{figure*}
    \centering
    \textbf{Medium Expert} \hspace{6.5cm} \textbf{Good Expert} \\
    \hspace{0.5cm}\textbf{Return} \hspace{3cm} \textbf{Regret} \hspace{3.5cm} \textbf{Return} \hspace{3cm} \textbf{Regret} \\
    \vspace{0.1cm}
    \includegraphics[width=0.24\linewidth]{figures_safety/freeway_medium_return}
    \includegraphics[width=0.24\linewidth]{figures_safety/freeway_medium_regret}
    \includegraphics[width=0.24\linewidth]{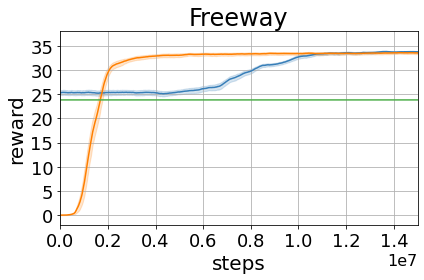}
    \includegraphics[width=0.24\linewidth]{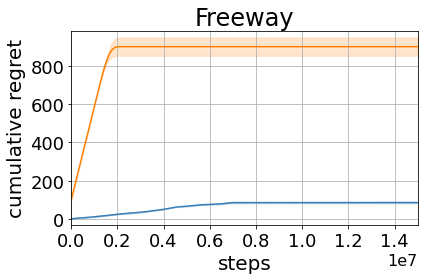}\\
    \includegraphics[width=0.24\linewidth]{figures_safety/breakout_medium_return}
    \includegraphics[width=0.24\linewidth]{figures_safety/breakout_medium_regret}
    \includegraphics[width=0.24\linewidth]{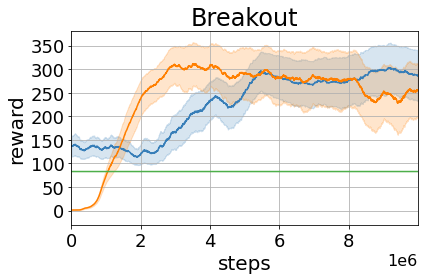}
    \includegraphics[width=0.24\linewidth]{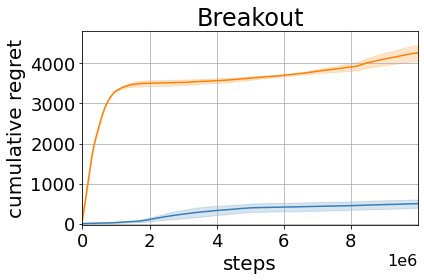}\\
    \includegraphics[width=0.24\linewidth]{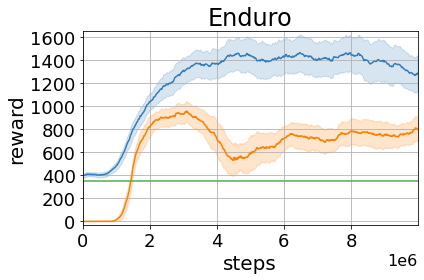}
    \includegraphics[width=0.24\linewidth]{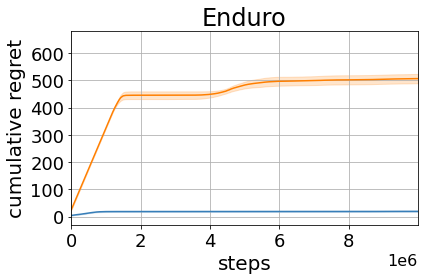}
    \includegraphics[width=0.24\linewidth]{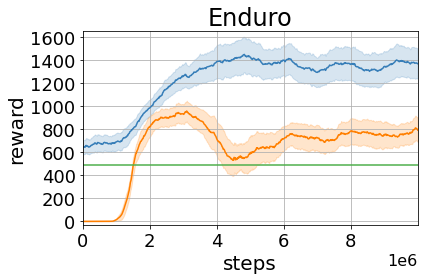}
    \includegraphics[width=0.24\linewidth]{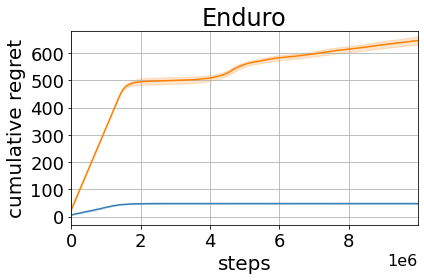}\\
    \includegraphics[width=0.24\linewidth]{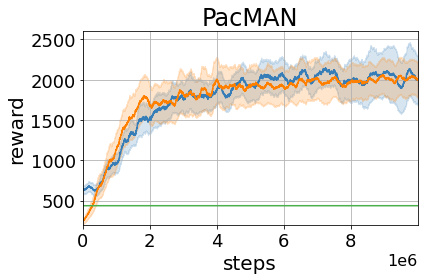}
    \includegraphics[width=0.24\linewidth]{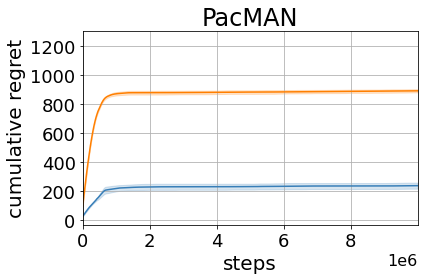}
    \includegraphics[width=0.24\linewidth]{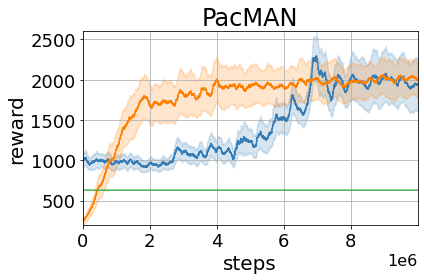}
    \includegraphics[width=0.24\linewidth]{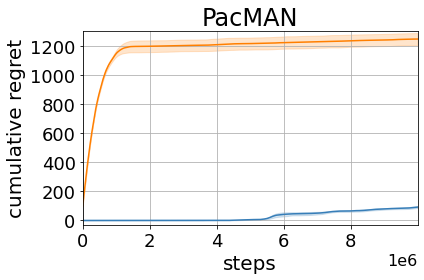}\\
    \includegraphics[width=0.24\linewidth]{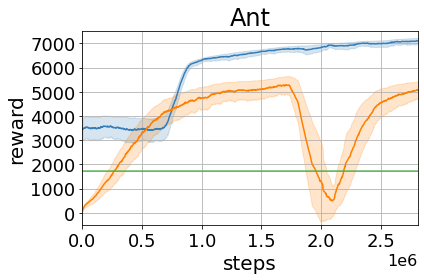}
    \includegraphics[width=0.24\linewidth]{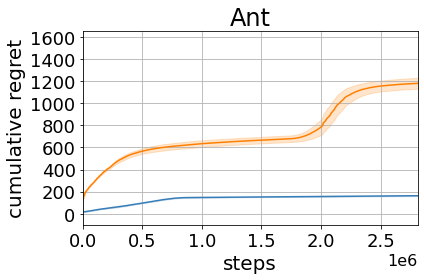}
    \includegraphics[width=0.24\linewidth]{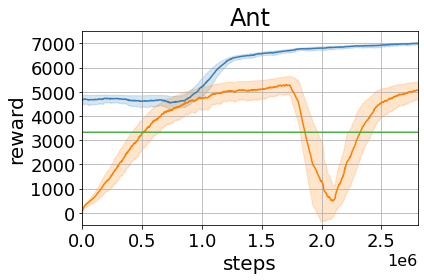}
    \includegraphics[width=0.24\linewidth]{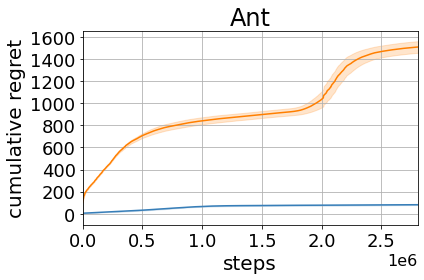}\\
    \includegraphics[width=0.24\linewidth]{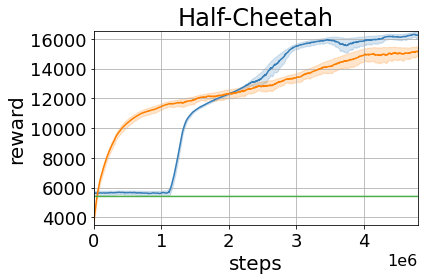}
    \includegraphics[width=0.24\linewidth]{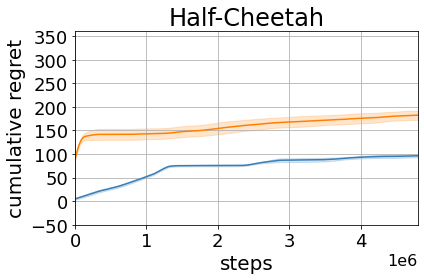}
    \includegraphics[width=0.24\linewidth]{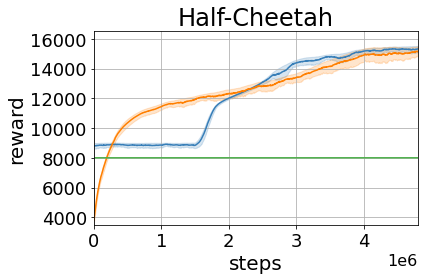}
    \includegraphics[width=0.24\linewidth]{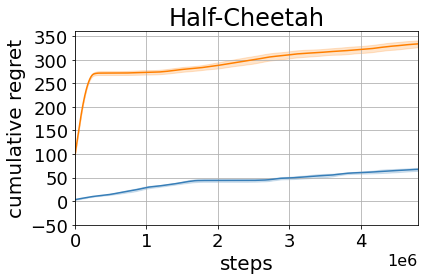}\\
    \includegraphics[width=0.24\linewidth]{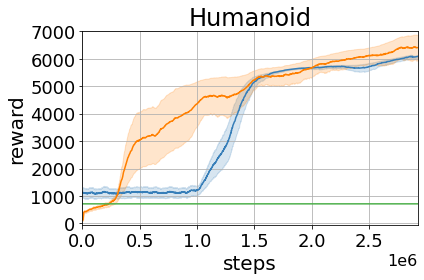}
    \includegraphics[width=0.24\linewidth]{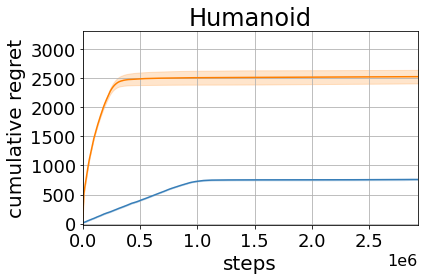}
    \includegraphics[width=0.24\linewidth]{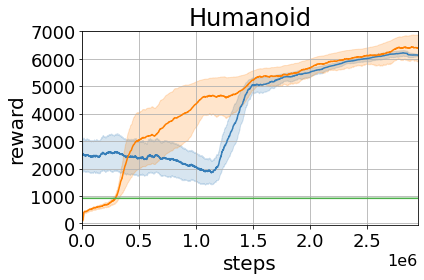}
    \includegraphics[width=0.24\linewidth]{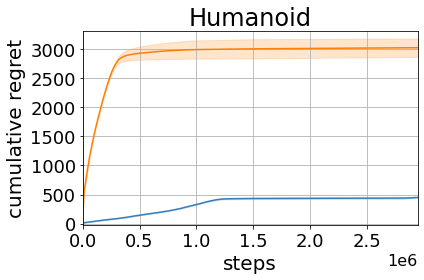}\\
    \includegraphics[width=0.5\linewidth,trim={0.1cm 0 0 0},clip]{figures_safety/legend}
    
    \caption{\textbf{External stability:} Performance and regret comparison between safe and oblivious learners. The performance of the benchmark is presented as the vertical green line. Opaque regions represent the standard deviation across 3 seeds. The goal of the safe agent throughout the entire learning process is, with high probability, to perform at least as good as the benchmark. The regret measures the accumulated sub-optimal regret, measured over entire episodes, with respect to the benchmark.}
    \label{fig: safe results}
\end{figure*}

\end{document}